\pdfoutput=1
\documentclass[journal]{IEEEtran}
\usepackage{cite}

\usepackage[cmex10]{amsmath}

\usepackage{array}
\usepackage{mdwmath}
\usepackage{mdwtab}

\usepackage{siunitx}
\usepackage{multirow}
\usepackage{bm}
\usepackage{mathtools}
\usepackage{amssymb}
\usepackage[linesnumbered,ruled,vlined]{algorithm2e}
\usepackage{booktabs}
\usepackage{hhline}
\usepackage{graphicx}
\usepackage{subcaption}
\usepackage{tabularx}

\begin{document}
\title{TROVE Feature Detection for Online Pose Recovery by Binocular Cameras}

\author{Yuance~Liu
        and~Michael~Z.~Q.~Chen*
\thanks{Yuance~Liu is with the Department
of Mechanical Engineering, The University of Hong Kong, Hong Kong, China (e-mail: aaarthur@hku.hk).}
\thanks{Michael~Z.~Q.~Chen (*, Corresponding Author) is with the School of Automation, Nanjing University of Science and Technology, Nanjing 210094, China (e-mail: mzqchen@outlook.com).}
}

\maketitle

\begin{abstract}
This paper proposes a new and efficient method to estimate 6-DoF ego-states: attitudes and positions in real time. The proposed method extract information of ego-states by observing a feature called ``TROVE'' (Three Rays and One VErtex). TROVE features are projected from structures that are ubiquitous on man-made constructions and objects. The proposed method does not search for conventional corner-type features nor use Perspective-n-Point (PnP) methods, and it achieves a real-time estimation of attitudes and positions up to 60 Hz. The accuracy of attitude estimates can reach 0.3 degrees and that of position estimates can reach 2 cm in an indoor environment. The result shows a promising approach for unmanned robots to localize in an environment that is rich in man-made structures.
\end{abstract}


\section{Introduction}
\IEEEPARstart{L}{ocalization} has always been pivotal for robots to autonomously navigate. Position in the Earth frame can be acquired by GPS sensors. Further fusing GPS information with measurements from other sensors, for instance: LiDAR, RGB-D cameras, ultrasonic sensors and etc., autonomous ground vehicles have already achieved robust and high-accuracy localization. However, for the limited payload and power supply some robots such as unmanned aerial vehicles (UAVs) can hardly afford to mount those powerful sensors and computers to obtain and process data. The blocking and reflecting of GPS signals from high-rises also impede accurate localization, let alone the indoor environment where GPS signal is often completely denied. Were UAVs to accurately navigate amid high-rises, complementary sensors other than GPS should be adopted. In that scenario, RGB cameras are the common option for UAVs to localize themselves because visual sensors can provide rich information at a cheap cost. The state-of-the-art visual SLAM (simultaneously-localization-and-mapping) algorithm has made significant improvement over the last decade. Methods such as LSD-SLAM \cite{engel2015large}, ORB-SLAM2 \cite{mur2017orb} and S-PTAM \cite{pire2017s} have recently drawn much attention and shown robust and accurate results. It is needless to mention that how pivotal real-time capability is to a system with fast dynamics like a quadrotor. Compared with other commonly used feature detectors: SIFT \cite{lowe2004distinctive}, SURF \cite{bay2006surf} and FAST \cite{rosten2006machine}, the aforementioned methods are much more efficient. The feature extraction and localization time for the SLAM algorithms can be roughly regarded as the tracking phase in a whole SLAM process. For LSD-SLAM, ORB-SLAM2 and S-PTAM, the tracking time using a rectified stereo image at a resolution of $1224\times370$ (can slightly vary by different rectification process) is 61.0, 49.5 and $\SI{55.2}{\milli\second}$ \cite{lowe2004distinctive, bay2006surf, rosten2006machine}, respectively. The proposed method in this paper completes that phase in only $\SI{2.8}{\milli\second}$ for an unrectified stereo image at a resolution of $1280\times720$. This allows the algorithm to run online at $\SI{60}{\hertz}$ in the experiment.

In this paper, a new method for attitude estimation and localization based on stereo vision is proposed. The method is dedicated to an environment where horizontal and vertical edges are common. Unlike classic PnP problems as described in \cite{horn1988closed, haralick1991analysis, lepetit2009epnp}, the proposed algorithm does not require any priori knowledge about the dimensions or positions of the observed object. It requires the included angle between two horizontal edges to be known and the existence of a vertex intersected by two horizontal edges and one vertical edge. This structure will be projected onto an image as three rays and one vertex. In this paper, such a feature is called a ``TROVE'' (Three Rays and One VErtex) feature and such a structure is called a ``TROVE'' structure. An environment with some priori knowledge is regarded as a semi-known environment. Some approaches have been attempted for pose estimates in such an environment, for instance \cite{claus2005reliable, fiala2010designing, bergamasco2016accurate, munoz2018mapping}. Those methods utilize preset markers or patterns, whereas the approach in this paper does not require any modifications on the environment. TROVE structures are ubiquitous in artificial constructions such as the outlines of buildings and corners of rooms and corridors. In the most common type of the structures, the horizontal edges are perpendicular to each other. The edges are usually formed by mutually orthogonal faces like ceilings, walls and facades of buildings. One of the earliest works utilizing those structures to extract ego-states can be found in \cite{coughlan1999manhattan}, where it referred those edges as Manhattan Grid and the environment rich in those structures as Manhattan World. In \cite{denis2008efficient, furukawa2009manhattan}, the authors utilized Manhattan Grid for attitude estimation and environment mapping. Some scholars have explored the possibility of estimating ego-states from other reliable world reference, including methods of horizon detection \cite{dusha2007attitude} and vanishing points \cite{demonceaux2007uav}. More variations of them can be found in a comprehensive survey in \cite{shabayek2012vision}. Nevertheless, localization has not been achieved in those methods.

The main \textit{contributions} of this paper are:
\begin{enumerate}
  \item A new feature has been proposed, which can associate its properties with the physical properties of the corresponding entity in the real world. Solution uniqueness of estimating ego-states from a feature is proved for the majority of cases, while easy-to-implement methods to discard incorrect solutions are also given.
  \item The solution is given in a closed-form expression. Unlike common tools in projection geometry \cite{hartley2003multiple} that need to solve eigenvectors of matrices or zeros of polynomials \cite{dhome1989determination} to estimate ego-states, the proposed method derives a direct closed-form expression of the estimates which greatly eases the computation burden.
  \item It has been demonstrated in the experiment that the algorithm can run in real time at $\SI{60}{\hertz}$ for $1280\times720$ stereo videos. The limiting factor for the efficiency is the camera's capability of capturing and transmitting images.
  \item The attitude estimates have high accuracy up to $\SI{0.2}{\degree}$. When fused with measurement from inertial measurement units (IMUs) in a straightforward way, the image estimates noticeably improve the conventional method.
  \item The method exhibits high accuracy in localization up to $\SI{2}{\centi\metre}$ on average.
\end{enumerate}
Edge detection in Manhattan-world scenario has been discussed by \cite{denis2008efficient, bazin2012globally}. Such a process is not the scope of this paper and is experimentally simplified by color segmentation.

The rest of this paper is organized as follows. Section II elaborates the definitions and the proof of the TROVE feature and its properties. Section III discusses the approach to detect edges and vertices of TROVE features, thus estimating 6-DoF ego-states. Section IV describes the experiments that evaluates the proposed method. All the findings and future work are summarized in Section V.

\section{TROVE Feature Detection}

In this section, the process to extract TROVE features in an image is presented. Such a feature is projected from a common structure, referred to as a TROVE structure, on man-made buildings both indoors and outdoors. Subsection \ref{subTROVEFeature} elaborates on the definition and properties of a standard TROVE feature with a real-world example. How a general TROVE feature could be transformed into a standard one and the proof of its properties is given in the latter half of this subsection. The following Subsections \ref{subEdgeDetection} and \ref{subVertexDetection} describe how a raw image is processed to extract the rays and the vertex of a TROVE feature in the experiment.

\subsection{TROVE Feature}\label{subTROVEFeature}

A TROVE feature consists of three rays and one vertex which the rays all originate from. A TROVE feature is projected from a TROVE structure, which consists of three edges and one vertex. All three edges of a TROVE structure intersect at the same vertex with one being vertical and the other two being horizontal in the Earth frame. Horizontal and vertical edges are ubiquitous on man-made structures, or more generally in Manhattan World \cite{coughlan1999manhattan}. The vertices, usually corners of room, corridors or outlines of buildings, man-made objects are naturally reliable landmarks for unmanned robots to refer to in unknown 3D environment. When those robots navigate indoors, the relative position to the environment is often much more important than their absolute position in the Earth frame. Those features are usually distinctively away from each other in distance and have properties such as orientations, positions, included angles by horizontal edges and being formed by different number of internal and external surfaces. More interestingly, these properties are associated with their physical properties such as shapes, relative distances and heights to the observer. These properties potentially enable them to have unique descriptors to be identified.

\begin{figure}
     \centering
     \begin{subfigure}[b]{0.3\linewidth}
         \centering
         \includegraphics[width=\linewidth]{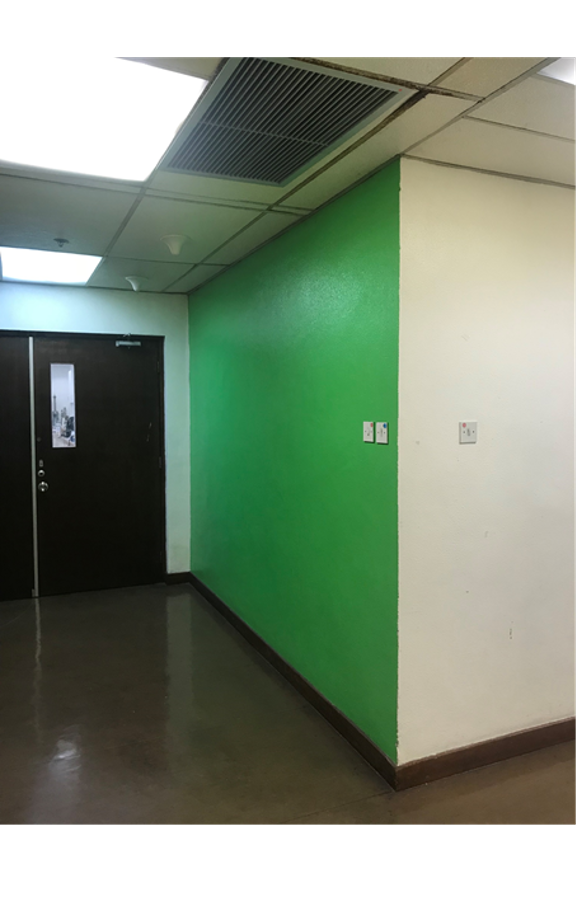}
         \caption{}
         \label{CorridorCornerSequence1}
     \end{subfigure}%
     \hfill
     \begin{subfigure}[b]{0.3\linewidth}
         \centering
         \includegraphics[width=\linewidth]{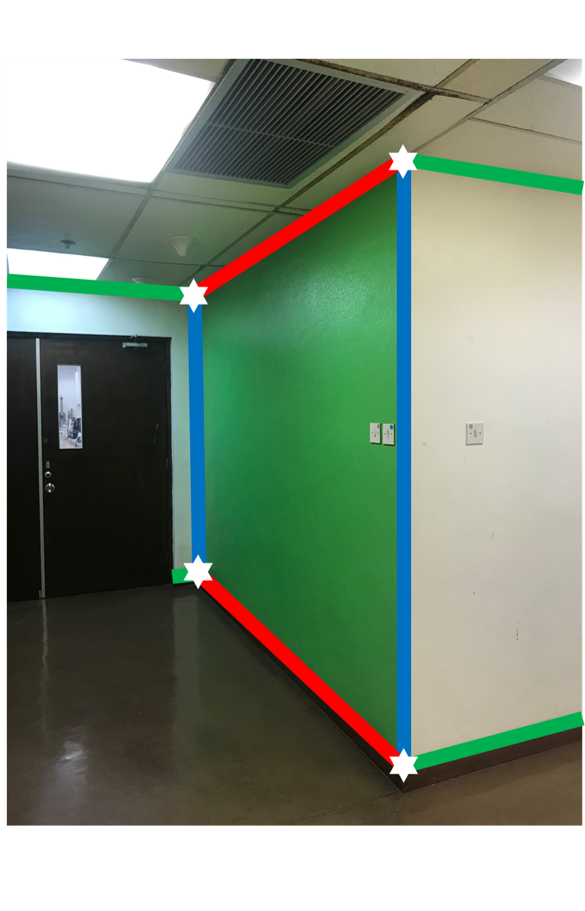}
         \caption{}
         \label{CorridorCornerSequence2}
     \end{subfigure}%
     \hfill
     \begin{subfigure}[b]{0.3\linewidth}
         \centering
         \includegraphics[width=\linewidth]{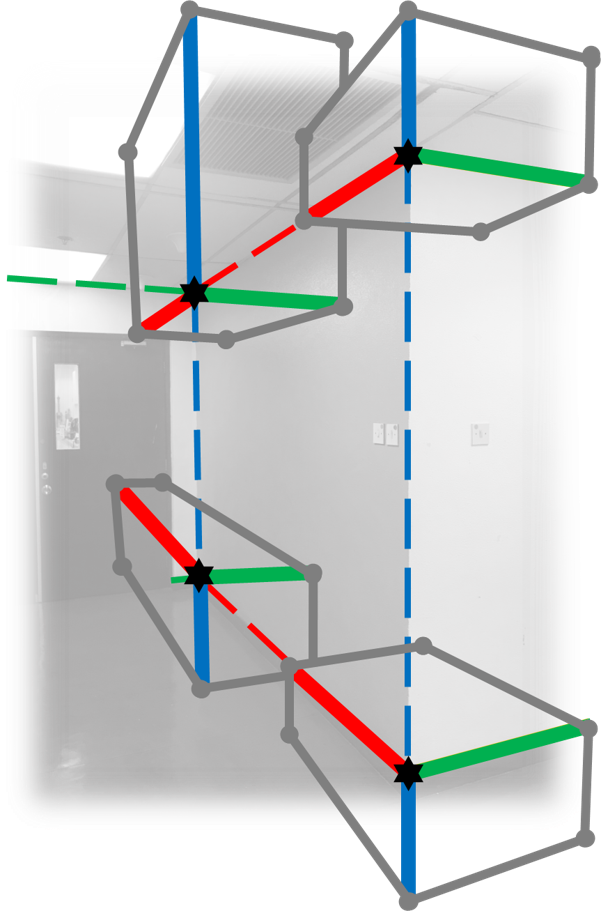}
         \caption{}
         \label{CorridorCornerSequence3}
     \end{subfigure}
     \caption{TROVE features in an image of a corridor. (a) an image of a corridor; (b) detected TROVE features, where the vertex is marked by a star and the edges by solid lines; (c) imaginary boxes constructed from features, where dashed lines are detected lines, color solid lines are edges of boxes on the detected lines and gray thin lines are the other edges of boxes.}
     \label{CorridorCornerSequence}
\end{figure}

A TROVE structure can be represented by a corner of an imaginary box. Fig. \ref{CorridorCornerSequence} shows an example of such a process in an image of a corridor. Fig. \ref{CorridorCornerSequence}(a) is the raw image taken in a corridor. Fig. \ref{CorridorCornerSequence}(b) shows the edges (solid lines) and vertices (white stars) of the TROVE features, where one vertical line and two horizontal lines intersect at a real physical vertex. If one imagines the walls, ceilings and floors of the corridor as the faces of boxes, each of the left two vertices is the intersection of three internal surfaces of a box, all visible to the camera. Each of the right two vertices is the intersection of three external surfaces of a box, two of which are visible and vertical and one of which is invisible and horizontal. In Fig. \ref{CorridorCornerSequence}(c), four imaginary boxes are constructed with the edges being collinear with the rays of the corresponding TROVE features in the image.

\subsubsection{Definition}

The definitions of the imaginary box are: 1) the box is a parallelepiped, i.e. a hexahedron with all faces as parallelograms; 2) of the three unparallel edges, one is vertical and the others are horizontal; 3) it has three faces visible to the camera and the intersections of them (three edges) are projected onto the image, collinear with the rays of the feature; 4) all edges are in equal unit length because dimensions are not concerned. Note that a necessary condition for a face crossing the principal point to be visible is that its outward normal vector points to the focal point.

The benefits of imagining a TROVE structure belonging an imaginary box are: 1) the positions/directions of vertices and edges remain the same; 2) the object frame can be uniquely defined by the box; 3) if more than one feature is found in an image, the model of such a hexahedron gives us a unique solution to attitude and position estimates when interpreted from a TROVE feature. Therefore, those boxes are reliable references for estimating ego-states of attitudes and positions.

\subsubsection{Standard Model}\label{subsectionStandardModel}

\begin{figure}[!htb]
    \centering
    \includegraphics[width=0.4\linewidth]{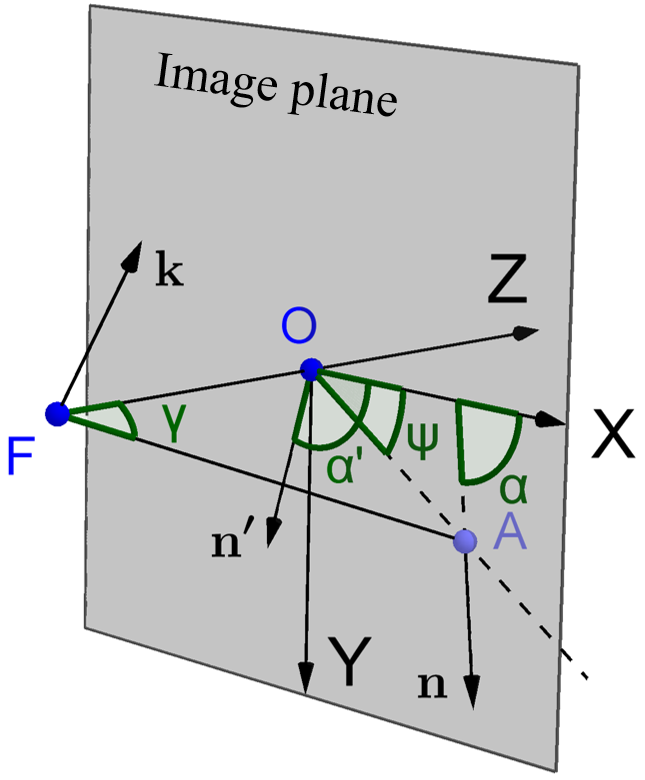}
    \caption{A projection is rotated so that the re-projected line passes through the principal point.}
    \label{rotatedAngle}
\end{figure}

In a standard model the vertex is projected onto the principal point. If the vertex is not originally on the principal point in the raw image, one can always rotate the observed object around the fixed focal point so that the new vertex can be projected onto the principal point. After such a rotation, the projection of edges has to adjust accordingly. Unless specifically mentioned, a TROVE feature is always rotated so that the vertex is projected onto the principal point. The following are the details of the transformation. Denote the intrinsic matrix of a pin-hole model camera as:
\begin{equation}
\bm{K} =
  \begin{bmatrix*}[c]
    f/\rho  & 0         & u_0   & 0\\
    0       & f/\rho    & v_0   & 0\\
    0       & 0         & 1     & 0
  \end{bmatrix*},
\end{equation}
where $f$ is the focal length and $\rho$ is the width of a pixel assuming $f$ and $\rho$ is identical in the x- and y-direction. $u_0$ and $v_0$ are the coordinate of the principal point in the image frame. In this paper, the image frame is defined that $u_0$ and $v_0$ are both $0$. As shown in Fig. \ref{rotatedAngle}, the projection of a line is represented by a unit vector $\bm{n}$ starting from $A$. $F$ is the focal point and $O$ is the principal point. If one rotates the object so that $A$ is projected onto $O$, the direct path is to rotate around the axis $\bm{k}$ by $\gamma$, where $\bm{k}$ is the normalized cross product of $\overrightarrow{FA}\times\overrightarrow{FO}$ passing through $F$ and $\gamma=\arctan{|OA|/|OF|}$. Let the position of $A$ be $(u,v)$, then one has:
\begin{align}
\bm{k}\!=\!\bigg(\frac{v}{\sqrt{v^2\!+\!u^2}},\frac{-u}{\sqrt{v^2\!+\!u^2}},0\bigg), \gamma\!=\arctan\!\bigg(\frac{\sqrt{u^2\!+\!v^2}}{f}\bigg). \label{gammaSolu}
\end{align}
Let $\bm{m}$ (not shown in the figure) be the rotated $\bm{n}$, $\bm{n}'$ be the projection of $\bm{m}$ in the image plane, $\psi$ be the angle that $\overrightarrow{OA}$ makes with the x-axis , $\alpha$ be the angle that $\bm{n}$ makes with the x-axis, $\alpha'$ be the angle that $\bm{n}'$ makes with the x-axis. $\alpha$ and $\alpha'$ are the inclination angles before and after rotation. It can easily be derived that $\bm{n}=[\cos{\alpha}, \sin{\alpha}, 0], \bm{k}=[\sin{\psi}, -\cos{\psi}, 0]$.

The objective is to find $\alpha'$. Rodrigues' rotation formula yields:
\begin{align}
\bm{m}=&\bm{n}\cos{\gamma}+(\bm{k}\times\bm{n})\sin{\gamma}+\bm{k}(\bm{k}\cdot\bm{n})(1-\cos{\gamma})\nonumber\\
    =&\begin{bmatrix*}[c]
        \cos{\gamma}\cos(\alpha-\psi)\cos{\psi}-\sin{\psi}\sin(\alpha-\psi)\\
        \cos{\gamma}\cos(\alpha-\psi)\cos{\psi}+\cos{\psi}\sin(\alpha-\psi)\\
        \cos(\alpha-\psi)\sin{\gamma}
      \end{bmatrix*},\label{mExpression}
\end{align}
Denote the $i^{\textnormal{th}}$ element of a vector $\bm{v}$ as $\bm{v}_i$. Note that: $\tan{\alpha'}=\bm{n}_2/\bm{n}_1=\bm{m}_2/\bm{m}_1$. By \eqref{mExpression} and difference formulas for tangent, one has:
\begin{align}
&\tan(\alpha'-\psi)=\tan(\alpha-\psi)/\cos{\gamma}\\
&\Rightarrow\alpha'=\psi+\arctan(\tan(\alpha-\psi)/\cos{\gamma}), \label{reprojectSolu}
\end{align}
where $\psi=\arctan(v/u)$. Thus, the re-projected line after rotation is found.

\begin{figure}
     \centering
     \begin{subfigure}[b]{0.25\linewidth}
         \centering
         \includegraphics[width=\linewidth]{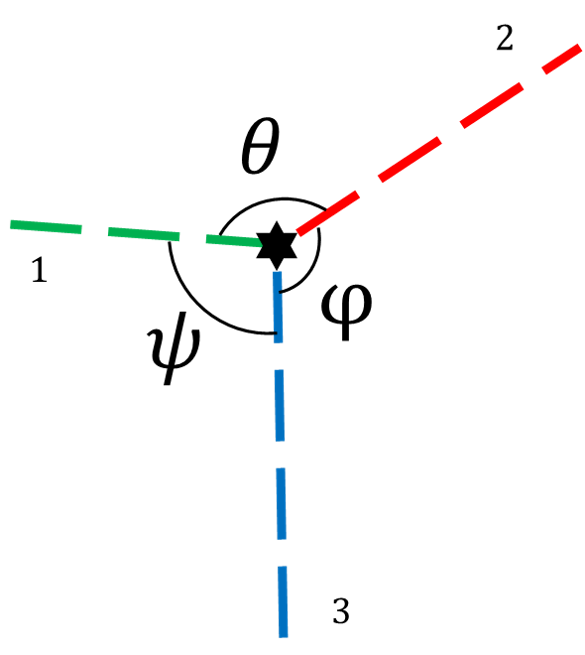}
         \caption{}
         \label{FeatureInterpretation1}
     \end{subfigure}%
     \hspace{0.03\linewidth}
     \begin{subfigure}[b]{0.25\linewidth}
         \centering
         \includegraphics[width=\linewidth]{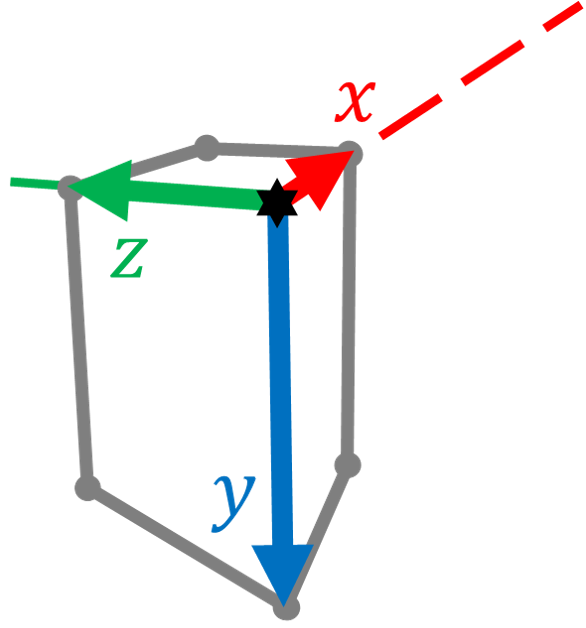}
         \caption{}
         \label{FeatureInterpretation2}
     \end{subfigure}%
     \hspace{0.03\linewidth}
     \begin{subfigure}[b]{0.25\linewidth}
         \centering
         \includegraphics[width=\linewidth]{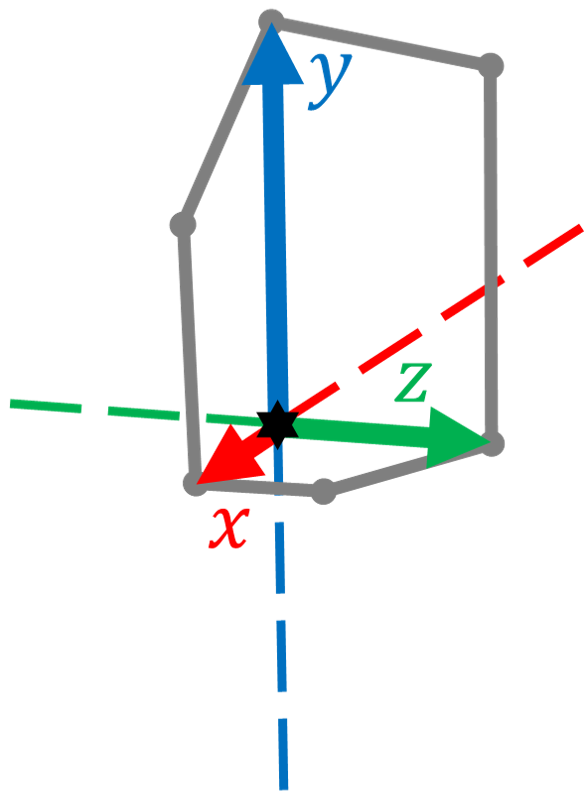}
         \caption{}
         \label{FeatureInterpretation3}
     \end{subfigure}

     \centering
     \begin{subfigure}[b]{0.25\linewidth}
         \centering
         \includegraphics[width=\linewidth]{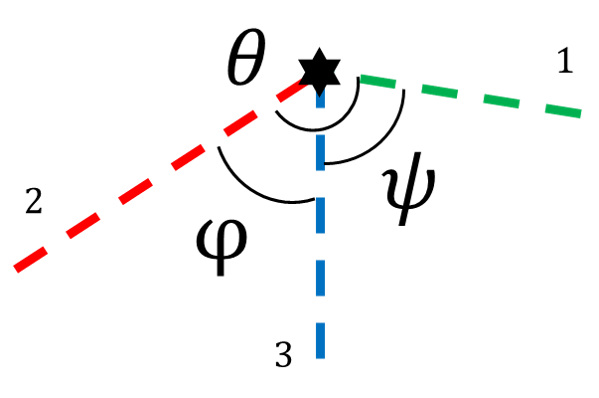}
         \caption{}
         \label{FeatureInterpretation4}
     \end{subfigure}%
     \hspace{0.03\linewidth}
     \begin{subfigure}[b]{0.25\linewidth}
         \centering
         \includegraphics[width=\linewidth]{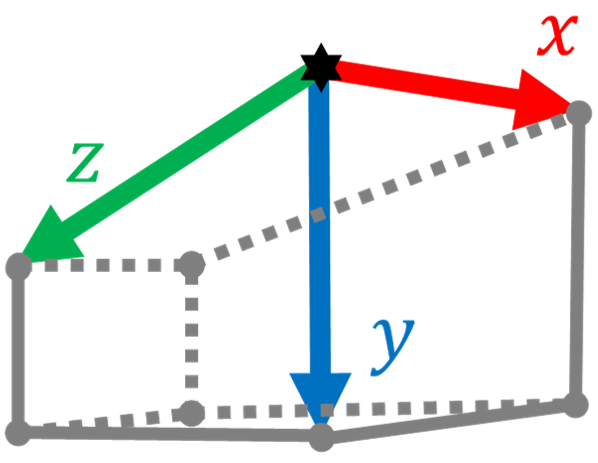}
         \caption{}
         \label{FeatureInterpretation5}
     \end{subfigure}%
     \hspace{0.03\linewidth}
     \begin{subfigure}[b]{0.25\linewidth}
         \centering
         \includegraphics[width=\linewidth]{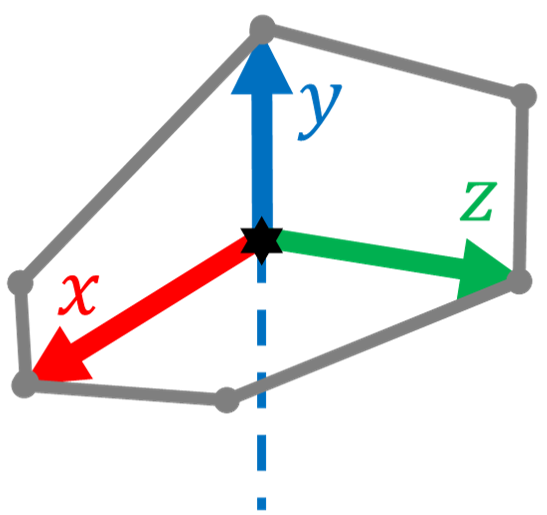}
         \caption{}
         \label{FeatureInterpretation6}
     \end{subfigure}
     \caption{Different interpretations from two types of features. (a) A detected TROVE feature that $\theta+\varphi+\psi=2\pi$; (b) and (c) are the two feasible interpretations from (a); (d) A detected TROVE feature that $\theta+\varphi+\psi<2\pi$; (e) is an infeasible interpretation of (d); (f) is a feasible interpretation of (d).}
     \label{FeatureInterpretation}
\end{figure}

After the vertex is rotated onto the principal point, the configuration ambiguity has to be tackled. It can be observed from Fig. \ref{CorridorCornerSequence} that the imaginary boxes are not constructed following the same rule. Whenever a TROVE feature is detected, one obtains three rays and their intersections in the image. How the box can be constructed from the detected rays is not unique. Suppose a TROVE feature is detected as shown in Fig. \ref{FeatureInterpretation}(a) that the sum of the three included angles $\theta, \varphi\textnormal{ and }\psi$ is $2\pi$. This type of feature is referred as the standard type. From a standard type of the feature in Fig. \ref{FeatureInterpretation}(a), the two and only two feasible imaginary boxes that can be interpreted are those in Fig. \ref{FeatureInterpretation}(b) and Fig. \ref{FeatureInterpretation}(c), so that 1) their edges are collinear with the rays; 2) their edges are either all on or all not on the rays; 3) the three faces formed by the three edges are visible. If $\theta+\varphi+\psi<2\pi$ as shown in Fig. \ref{FeatureInterpretation}(d), that TROVE feature belongs to a different type. In such a case, among the three faces that are formed by the three edges, one must be occluded as shown in Fig. \ref{FeatureInterpretation}(e). In this case, the ray that is not the side of the largest angle should be flipped so that the feature is transformed into a standard type. It is noteworthy that the two included angles of the new imaginary box with the flipped ray as their side will be the supplementary angles of the corresponding original ones. One feasible interpretation after the flipping is illustrated in Fig. \ref{FeatureInterpretation}(f). In the following discussion, all features are the standard type.

The objective frame is also uniquely defined for each imaginary box. The object frame is a right-hand Cartesian coordinate system with the origin at the box's vertex which is the intersection of the three visible faces. As shown in Fig. \ref{FeatureInterpretation}, the y-axis is defined to be collinear with the vertical edge of the box. The x-axis is defined to be collinear with a horizontal edge of the box. The box is then in the positive y-direction and positive z-direction of the object frame. Note that the included angle by the two horizontal edges can be either acute, right or obtuse. Therefore, the z-axis is not necessarily aligned with any edge of the box. Furthermore, the box is imaginary that its edges do not necessarily overlap with any physical edges.

\subsubsection{Properties}\label{subsubsecFeatureProperties}

The three rays of a TROVE feature have three included angles. Recall that the corresponding included angles in 3D space are either right between vertical and horizontal edges or a known angle between two horizontal edges, which is smaller than $\pi$ as a property of a hexahedron. Denote the angle between two horizontal edges as $\beta$ and the corresponding projected angle in the image as $\theta$. Denote the projection of two angels in the image, each made by one vertical and one horizontal edge, as $\varphi$ and $\psi$ respectively. In this paper, the direction of an edge means the direction that points from the vertex to the other end.

The properties can be summarized as: 1) $\beta>\theta$ in all cases; 2) if $\beta\in(0,\pi/2)$, $\theta\in(0,\pi)$ and $\varphi, \psi \in(\pi/2,\pi)$; 3) if $\beta=\pi/2$, $\beta,\varphi,\psi\in(\pi/2,\pi)$; 4) if $\beta\in(\pi/2,\pi)$, $\theta\in(\pi/2,\pi)$ and $\varphi, \psi\in(0,\pi)$ with at most one belongs to $(0,\pi/2)$. These properties are the premises of the uniqueness of the analytic solution in the following section. They can also be utilized to screen out incorrect detections, identify horizontal or vertical edges and estimate the orientation and relative height of the observed structure. The proof is given in the following paragraphs.

\begin{figure}[!htb]
    \centering
    \includegraphics[width=0.8\linewidth]{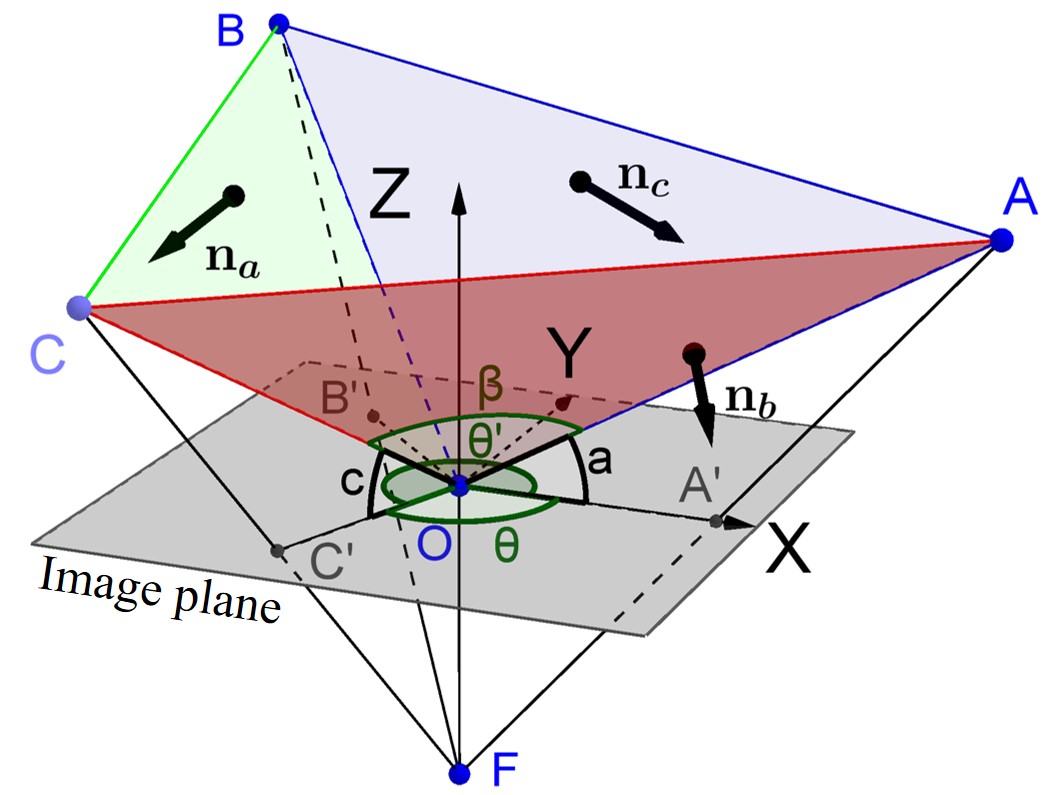}
    \caption{A structure is projected onto the image plane as a TROVE feature.}
    \label{IncludedAngleProjection}
\end{figure}

Definitions, notations and basic properties of the model are given before the proof of the properties. Since the projected vertex is at the principal point, the vertex of a TROVE structure can be translated onto the principal point for convenience without changing the projected angle. An illustration of a translated TROVE structure and its projection is shown in Fig. \ref{IncludedAngleProjection}, where three unit segments $OA$, $OB$ and $OC$ belong to the three edges of the TROVE structure. They are projected as $OA'$, $OB'$ and $OC'$, respectively. In the earth frame, $OA$ and $OC$ are horizontal while $OB$ is vertical. $\overrightarrow{OA},\overrightarrow{OB}$ follows the positive x- and y-direction of the object frame and subsequently $\overrightarrow{OB}$ follows the cross product $\overrightarrow{OC}\times\overrightarrow{OA}$. Without loss of generality, the TROVE structure is rotated around the z-axis so that $OA'$ is on the positive x-axis. $F$ is the focal point at the origin of the camera frame, while $O$ is at $(0,0,f)$. Denote $\angle AOA'$ as $a\in(-\pi/2,\pi/2)$ and $\angle COC'$ as $c\in(-\pi/2,\pi/2)$. Positive signs of $a,c$ mean that $\overrightarrow{OA},\overrightarrow{OC}$ point to the positive direction of z-axis. Note that the projection of an edge has to be observable, therefore $|a|\neq\pi/2, |c|\neq\pi/2$. Denote the angle between $OC'$ and the positive x-axis in a counterclockwise direction as $\theta'\in(0,2\pi)$, the angle between $OC'$ and $OB'$ as $\varphi$, the angle between $OA'$ and $OB'$ as $\psi$. $\varphi$ and $\psi$ are projected from the angles between the horizontal edges and the vertical edge of the TROVE structure. For clarity, $\varphi$ and $\psi$ are not annotated in the figure.

The position of $A$ in the camera frame can be found as ${}^{\{\textnormal{C}\}}\!\bm{P}_A=(\cos{a},0,\sin{a}+f)^T$ and $C$ in the camera frame as ${}^{\{\textnormal{C}\}}\!\bm{P}_C=(\cos{c}\cos\theta',\cos{c}\sin\theta',\sin{c}+f)^T$. The projected point $A'$ and $C'$ can be expressed in a homogeneous form in the image frame as ${}^{\{\textnormal{I}\}}\!\bm{\widetilde{P}_{A'}} =\bm{K}{}^{\{\textnormal{C}\}}\!\bm{P}_A$ and ${}^{\{\textnormal{I}\}}\!\bm{\widetilde{P}_{C'}} =\bm{K}{}^{\{\textnormal{C}\}}\!\bm{P}_C$. The vectors in the camera frame pointing to the same direction as $\overrightarrow{OA},\overrightarrow{OB}$ and $\overrightarrow{OC}$ are denoted as $\bm{a}$, $\bm{b}$ and $\bm{c}$, respectively. The corresponding projected vectors in the image frame $\overrightarrow{OA'}$, $\overrightarrow{OB'}$ and $\overrightarrow{OC'}$ are denoted as $\bm{a'}$, $\bm{b'}$ and $\bm{c'}$, respectively. Since vector magnitude does not change the concerned included angles, these vectors can be chosen as unit vectors following the original direction as:
\begin{align}
    \bm{a}=&(\cos{a}, 0, \sin{a}),\\
    \bm{b}=&(\sin{a}\cos{c}\sin\theta', \cos{a}\sin{c}-\nonumber\\
    &\sin{a}\cos{c}\cos\theta', -\cos{a}\cos{c}\sin{\theta'}),\\
    \bm{c}=&(\cos{c}\cos\theta', \cos{c}\sin\theta', \sin{c}),\\
    \bm{a'}=&(1, 0), \\
    \bm{b'}=&(\sin{a}\cos{c}\cos\theta', \cos{a}\sin{c}-\sin{a}\cos{c}\cos\theta'),\\
    \bm{c'}=&(\cos\theta', \sin\theta').
\end{align}

By dot product, it can be calculated that:
\begin{equation}\label{cosBeta}
\cos\beta=\cos{a}\cos{c}\cos\theta'+\sin{a}\sin{c}.
\end{equation}
It is noteworthy that $\theta+\theta' = 2\pi$ and $\cos\theta'=\cos\theta$. By cross product of $\bm{c}\times\bm{b}, \bm{a}\times\bm{c}\textnormal{ and }\bm{b}\times\bm{a}$, three vectors are obtained, which are denoted as $\bm{n}_a, \bm{n}_b\textnormal{ and }\bm{n}_c$ respectively. These three vectors are normal to the three faces of the imaginary box and pointing away from the box as shown in the figure. Denote the $i^{\textnormal{th}}$ element of a vector $\bm{v}$ as $\bm{v}_i$. Note that all faces are visible and cross the principal point. Hence, their normal vectors' z-components are negative:
\begin{align}
\bm{n}_{a3}<0&\Rightarrow\cos{a}\sin{c}\cos\theta'<\sin{a}\cos{c},\label{nzComponent1}\\
\bm{n}_{b3}<0&\Rightarrow\cos{a}\cos{c}\sin\theta'<0,\label{nzComponent2}\\
\bm{n}_{c3}<0&\Rightarrow\sin{a}\cos{c}\cos\theta'<\cos{a}\sin{c}.\label{nzComponent3}
\end{align}
Since $\bm{b}_3=-\bm{n}_{b3}>0$, the vertical edge can never point to the focal point. From \eqref{nzComponent2}, one can easily obtain $\theta<\pi$.

As the premise for further derivations, the authors first prove that $\theta>\beta$ in all cases.

Study the case of $ac\geq0$. If $\bm{a}_3<0, \bm{c}_3<0$, namely $\sin{a}<0, \sin{c}<0$. By multiplying the two sides of \eqref{nzComponent1} and \eqref{nzComponent3} one has $\cos^2\theta'>1$, which is a contradiction. Therefore, $a, c$ cannot both be negative, i.e. $\overrightarrow{OA},\overrightarrow{OC}$ cannot both point towards the focal point. One must have $a,c\in[0,\pi/2)$. Assume $\cos\beta\leq\cos\theta'$, namely $\beta>\theta$. Without loss of generality, suppose $c\geq a$. By the assumption, one has:
\begin{equation}\label{cosBetaLesCosTheta}
\sin{a}\sin{c}\leq(1-\cos{a}\cos{c})\cos\theta'.
\end{equation}
When $a=0$, substituting $a$ into \eqref{cosBetaLesCosTheta} and \eqref{nzComponent1} yields a contradiction. Thus, $a,c\neq0$. Substituting \eqref{nzComponent3} into \eqref{cosBetaLesCosTheta} gives:
\begin{align}
&\sin{a}\sin{c}<(1-\cos{a}\cos{c})\frac{\cos{a}\sin{c}}{\sin{a}\cos{c}}\\
&\implies \cos{a}<\cos{c},
\end{align}
which contradicts $c\geq a$ in the range of $(0,\pi/2)$. Therefore, the assumption must be false and $\beta<\theta$ when $ac\geq0$.

Study the case of $ac<0$. In this case one and only one angle can be negative. Without loss of generality, suppose $c\in(-\pi/2,0), a\in(0,\pi/2)$. Substituting $\cos\beta$ by \eqref{cosBeta}, $\cos\beta-\cos\theta'$ can be rewritten as:
\begin{equation}\label{CBetaMinusCTheta}
  (\cos{a}\cos{c}-1)\cos\theta'+\sin{a}\sin{c}.
\end{equation}
Since $\sin{c}<0$, \eqref{nzComponent1} and \eqref{nzComponent3} can be transformed into:
\begin{align}
&\frac{\sin{a}\cos{c}}{\cos{a}\sin{c}}<\cos\theta'<\frac{\cos{a}\sin{c}}{\sin{a}\cos{c}}\label{transformedTheta}<0\\
&\implies \cos^2{a}\sin^2{c}<\cos^2{c}\sin^2{a}.\label{cosSinLesCosSin}
\end{align}
Note $\cos{a}\cos{c}-1<0$. Substituting \eqref{transformedTheta} into \eqref{CBetaMinusCTheta} yields:
\begin{equation}\label{CBetaMinusCThetaTrans}
\cos\beta-\cos\theta' = \frac{\sin{a}(\cos{a}-\cos{c})}{\cos{a}\sin{c}}.
\end{equation}
By \eqref{cosSinLesCosSin}, one knows that $\cos{a}<\cos{c}$. Further note $\sin{c}<0$. \eqref{CBetaMinusCThetaTrans} must be greater than 0. Thus, $\beta<\theta$.

With the previous efforts, the properties of the feature with respect to the properties of the observed structure can also be derived. The properties are discussed regarding the range of $\beta$ in three cases. \\
\textbf{Case 1:} $\beta\in(0,\pi/2)$

First, $\bm{a}_3>0,\bm{c}_3>0$ is proved. Assume $\bm{a}_3\bm{c}_3\leq0$, namely $\sin{a}\sin{c}\leq0$. Note that from \eqref{cosBeta} $\cos{\beta}>0\Rightarrow\cos{\theta'}>0$ when $ac<0$. $\bm{n}_{a3}\bm{n}_{c3}$ gives:
\begin{align}
\bm{n}_{a3}\bm{n}_{c3} =&(\cos{a}\sin{c}\cos\theta'-\sin{a}\cos{c}\cos{c})\times\nonumber\\
&(\sin{a}\cos{a}\cos{c}\cos\theta'-\cos^2a\sin{c})\label{}\\
=&\cos{a}[\sin{a}\cos{a}\sin{c}\cos{c}(\cos^2\theta'+1)\nonumber\\
&-(\sin^2a\cos^2c+\cos^2a\sin^2c)\cos\theta'].\label{na3nc3}
\end{align}
Note it is always true that $\cos{a}>0,\cos{c}>0$. Obviously, the expression in \eqref{na3nc3} is less than 0, which contradicts the fact that $\bm{n}_{a3},\bm{n}_{c3}>0$. Therefore, the assumption $\bm{a}_3\bm{c}_3\leq0$ must be false. Since it was previously pointed out that $a, c$ cannot both be negative angles, $\bm{a}_3,\bm{b}_3,\bm{c}_3>0$.

Second, for any two orthogonal vectors $\bm{p},\bm{q}$ starting from the origin $O$ in 3D space, if $\bm{p}_3\bm{q}_3>0$ their projected included angle must belong to $(\pi/2,\pi)$ (the proof is omitted for brevity). Since $\theta>\beta$, it can be concluded that $\theta\in(0,\pi)>\beta$ and $\varphi,\psi\in(\pi/2,\pi)$.\\
\textbf{Case 2:} $\beta=\pi/2$

Similar to \textbf{Case 1}, it can be proved that $\bm{a}_3,\bm{b}_3,\bm{c}_3>0$. All the projected angles included by the mutually orthogonal edges must $\in(\pi/2,\pi)$.\\
\textbf{Case 3:} $\beta\in(\pi/2,\pi)$

Study the case where $ac\geq0$. For any two orthogonal vectors $\bm{p},\bm{q}$ starting from the origin $O$ in 3D space, if $\bm{p}_3\bm{q}_3=0$ their projected included angle must be $\pi/2$ (the proof is omitted for brevity). Hence,  one of $\varphi$ and $\psi$ will be a right angle. If $ac<0$, it is obvious that $\cos\theta'<0$ from \eqref{na3nc3} $>0$. If $\bm{p}_3\bm{q}_3<0$ their projected included angle must belong to $(0,\pi/2)$ (the proof is omitted for brevity). Hence, one and only one of $\varphi$ and $\psi$ will be an acute angle. In summary, one of the following statements will be true: 1) $\theta>\beta>\pi/2$ and $\varphi,\psi\in[\pi/2,\pi)$ with at most one of $\varphi$ and $\psi$ equals to $\pi/2$; 2) $\theta>\beta>\pi/2$ and one of $\varphi$ and $\psi$ belongs to $(0,\pi/2)$ with the other belonging to $(\pi/2,\pi)$.

\subsection{Edge Detection}\label{subEdgeDetection}
\begin{figure}[!htb]
  \centering
  \includegraphics[width = 2 in]{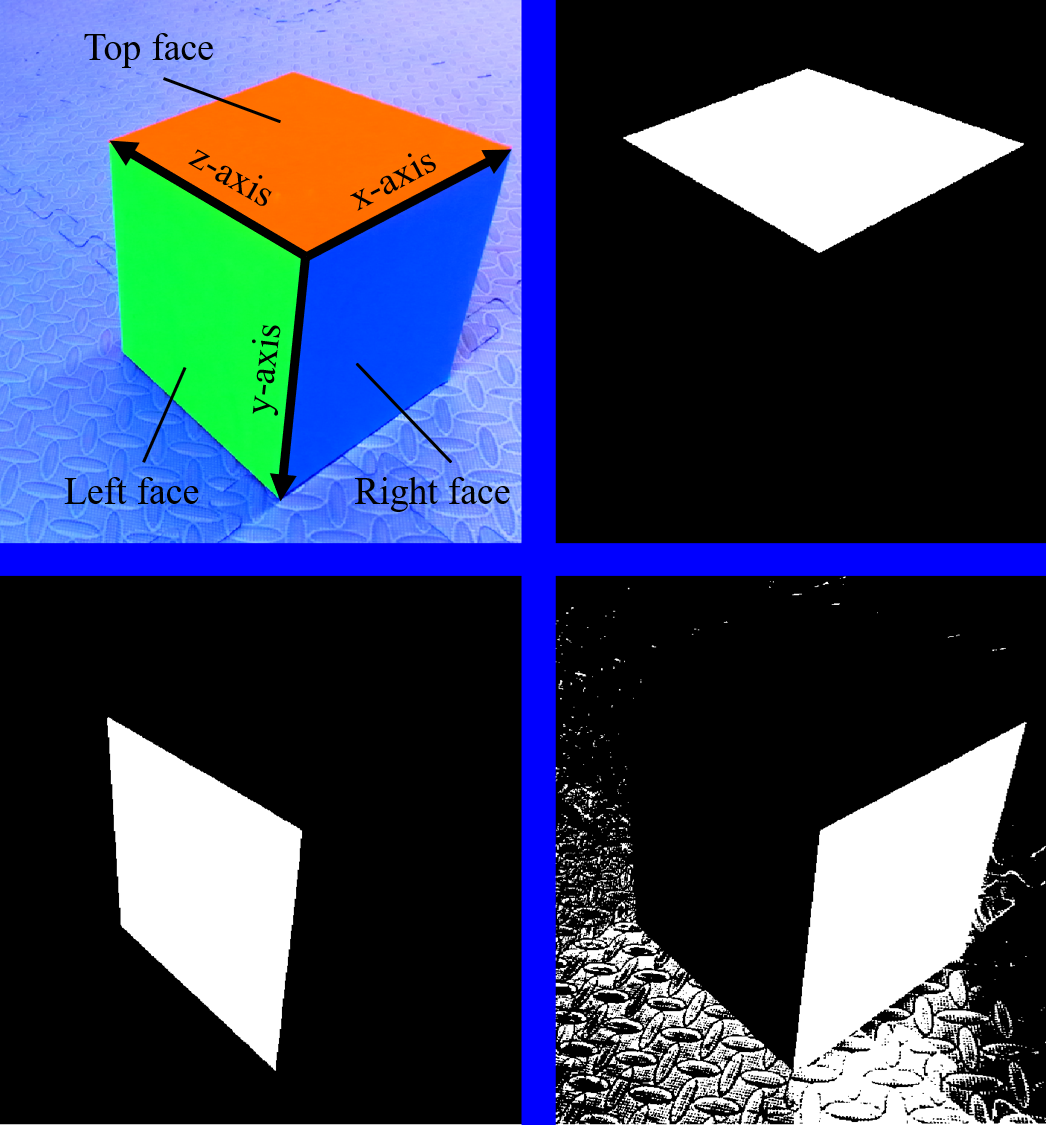}
  \caption{Top left: the original image with notation of each face and axis of the object frame; top right: pixels labeled red; bottom left: pixels labeled green; bottom right: pixels labeled blue.}
  \label{ColorPatchCombineText}
\end{figure}

Three adjacent faces of a cuboid are colored in red, green and blue, respectively referred to as the top, left and right faces. Multi-thresholding is applied for color segmentation, concept of which was introduced in \cite{kurugollu2001color}. The standards of labeling a pixel are chromaticity, the proportion of each RGB value, and intensity, the absolute value of each primary color. The pixels at the edges are often labeled as none of the target colors, where one color transits to another. An example of recognized patches and the x-axis, y-axis, z-axis of the object frame are shown in Fig.~\ref{ColorPatchCombineText}.

\begin{figure}[!htb]
  \centering
  \includegraphics[width = 8 cm]{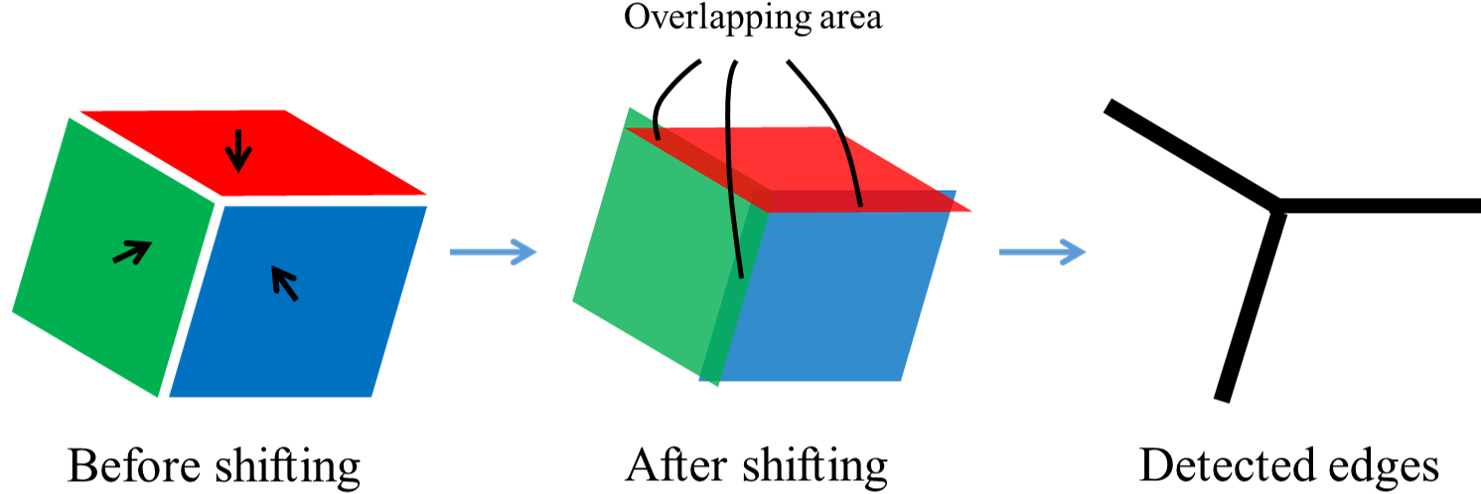}
  \caption{Shifted color patches to find edges \cite{liu2017novel}}
  \label{ShiftDetectingEdge}
\end{figure}

The edges are found at the pixels that neighbor more than one type of color pixels. One can find the edge by examining every pixel whether it has neighbors of two types of color pixels. Direct implementation of this logic turns out to be computationally intensive since one needs to navigate through all the pixels and examine all their neighbors. Specifically with a search area of a $3\times3$ window, to determine if a pixel belongs to one of the three edges all its 8 neighboring pixels have to be examined. For a $1920\times1080$ image, it requires about 50 million operations, let alone a $4\times4$ search window. Even for a $1280\times720$ image, it still requires about 22 million operations, which is demanding for a processing unit especially on payload-limited robots such as UAVs. The process can be optimized by shifting color patches \cite{liu2017novel}. As shown in Fig.~\ref{ShiftDetectingEdge}, the patches are shifted towards the gap. Then the candidate points for the edge detection are found in the overlapping area.

However, there exists a contradiction that one needs to know the direction of a gap before the gap is even found. To solve this dilemma, one can make use of priori estimates of attitudes, either from itself or other sensors. If the camera frame's upward direction also points upward in the Earth frame, the top face of the cuboid is on the upper side compared with the other two faces. To detect the projection of the y-axis edge, the top face is shifted downward and leftward. Overlapping area of those two shifted patches are the cluster of the candidate pixels. Similar methods apply to detect the other two edges and also the case where the camera is upside down.

Different from \cite{liu2017novel}, this propose method is not realized by copying and writing pixel values into a new position but optimized by examining the pixel at the position where it is supposed to be shifted from. Pseudo code of this process, taking the case when the camera is not upside down, is illustrated in Alg. \ref{GeneralAlgo}. It is noteworthy that the algorithm takes unrectified images and only rectifies the coordinates of the recognized edge pixels. Specifically for a pixel $A$, if it belongs to none of the target faces or the top face the program skips to the next pixel. If $A$ belongs to the right face, one examines the pixel $B$ to the left of the current $A$ whose distance to $A$ is defined by a search width $W$. If $B$ belongs to the left face, the middle pixel $C$ between $A$ and $B$ belongs to the y-axis. The coordinate of $C$ is rectified according to stereo parameters and then stored for line detection. Then the program continues to the next pixel. Similar methods apply to the other edges. Note that at most three operations are performed for each pixel and the overall number of operations for a $1920\times1080$ image has been reduced from 50 million to less than 3 million on average. Having the candidate coordinates for each edge, the authors use the Random Sample Consensus (RANSAC) method \cite{fischler1981random} to find a line from a cluster of points. In image processing, the method of Hough Transform is also commonly adopted to detect lines through points. But this is an exhaustive method that calculates all possible lines through every point. In this scenario RANSAC is far more efficient, especially under the condition that there exist many outliers. In the experiment, RANSAC is more than 40 times faster than the Hough Transform method.

\begin{algorithm}
    \KwData{Unrectified images of an observed object}
    \KwResult{TROVE feature lines in the image}
    Read an image matrix $\bm{M}_I$ from the camera\;
    \tcc{Label each pixel as top, left, right or none of the target surfaces. }
    Create a label matrix $\bm{M}_L$\;
    \For{All pixels $p_{i,j}$ in $\bm{M}_I$}
        {$\bm{M}_L(i,j)$ = colorSegmentation($p_{i,j}$)\;}
    \tcc{Identify edge pixels. }
    Define a search width $W$\;
    \For{All entries $l_{i,j}$ in $\bm{M}_L$}
    {
        \Switch{the label of $l_{i,j}$}
        {
            \Case{none of the target faces}
                {Go to the next entry\;}
            \Case{top face}
                {Go to the next entry\;}
            \Case{right face}
            {
                \If{$l_{i,j-W}$ is the left face}
                    {Rectify coordinate $(i, j-W/2)$\; Add the coordinate to $zAxisCoor$\;}
                \ElseIf{$l_{i-W,j-W}$ is the top face}
                    {Rectify coordinate $(i-W/2, j-W/2)$\; Add the coordinate to $xAxisCoor$\;}
            }
            \Case{left face}
            {
                \If{$l_{i-W,j+W}$ is the top face}
                    {Rectify coordinate $(i-W/2, j+W/2)$\; Add the coordinate to $yAxisCoor$\;}
            }
        }
    }
    \tcc{Detect lines by RANSAC. }
    ($xAxis$, $yAxis$, $zAxis$) = RANSAC($xAxisCoor$,$yAxisCoor$,$zAxisCoor$)\;
    \caption{TROVE feature line detection}
    \label{GeneralAlgo}
\end{algorithm}

\subsection{Vertex Detection}\label{subVertexDetection}
The location of the vertex where three edges intersect should also be estimated. The necessity of vertex location information will be discussed in the following section. In practice, the three detected edges never intersect at exact one point, and thus an estimate is necessary. A line in 2D space passing through $(x,y)$ can be represented by $\rho=x \cos{\theta} + y \sin{\theta}$, where $\rho$ is the signed normal distance from the origin to the line; $\theta$ is the angle between the normal of the line and the positive $x$-axis in a range of $[-\pi/2,\pi/2)$.

In this paper, the vertex is found at the position where the sum of its squared distance to the three edges is minimized. Denote the position of a vertex as $\bm{V} = [x_o, y_o]^\intercal$. Let a line $i$ in the Hough Space be $(\rho_i, \theta_i)$. The signed distance $e_i$ from this line to the vertex can be represented by:
\begin{equation}\label{OneDistance}
  e_i = \rho_i - (x_o \cos{\theta_i} + y_o \sin{\theta_i}).
\end{equation}

Let $i=1,2,3$ denote the projection of the three edges of a TROVE structure, respectively. Denote:
\begin{equation}\label{VertexFindingDenotation}
  \bm{\rho} =\!
  \begin{bmatrix*}[c]
    \rho_1 \\
    \rho_2 \\
    \rho_3
  \end{bmatrix*}\!,
  \bm{R} =\!
  \begin{bmatrix*}[c]
    \cos{\theta_1} & \sin{\theta_1} \\
    \cos{\theta_2} & \sin{\theta_2} \\
    \cos{\theta_3} & \sin{\theta_3}
  \end{bmatrix*}\!,
  \bm{E} =\!
  \begin{bmatrix*}[c]
    e_1 \\
    e_2 \\
    e_3
  \end{bmatrix*}\!,
  \bm{V} =\!
  \begin{bmatrix*}[c]
    x_o \\
    y_o
  \end{bmatrix*}\!.
\end{equation}

Their distance to the vertex can be represented in a simple matrix form: $\bm{E} = \bm{\rho} - \bm{RV}$. The estimated position $\widehat{\bm{V}}$ is found by:
\begin{equation}\label{MinimumVertex}
  \widehat{\bm{V}}=\arg\min\bm{E}^\intercal\bm{E}\implies\bm{\widehat{V}}=(\bm{RR^{\intercal}})^{\textnormal{-1}}\bm{R^{\intercal}\rho}.
\end{equation}
Subsequently the sum of squared errors can also be obtained, which is utilized to screen out invalid vertices.

\section{Pose Estimation}

The previous section has discussed the properties of a TROVE feature associated with their counterpart: a TROVE structure in 3D space and the detection methods for the projected edges and the vertex. Knowing the edges and vertex, one can estimate the attitudes and positions of the camera in the object frame. The initial orientation is defined that the x-axis and y-axis of the camera frame are aligned with those of the object frame. The attitude is defined as the rotation from the initial orientation to the current one in the object frame. By comparing the vertex in two stereo images and integrating the attitude estimate, the position of the camera can be recovered.

When a line is projected onto an image, the line must be on the plane that passes through the focal point and the projected line. Once a TROVE feature is detected, three planes that the corresponding three rays lie on are also determined. For simplicity, the three edges of the imaginary box are denoted as x-, y- and z-edge. The x- and y-edge align with the x- and y-axis of the object frame, respectively while the z-edge is not necessarily aligned with the z-axis. Denote the plane where the x-edge lies as the x-plane, that where the y-edge lies as the y-plane and that where the z-edge lies as the z-plane.

\begin{figure}[!htb]
  \centering
  \includegraphics[width = \linewidth]{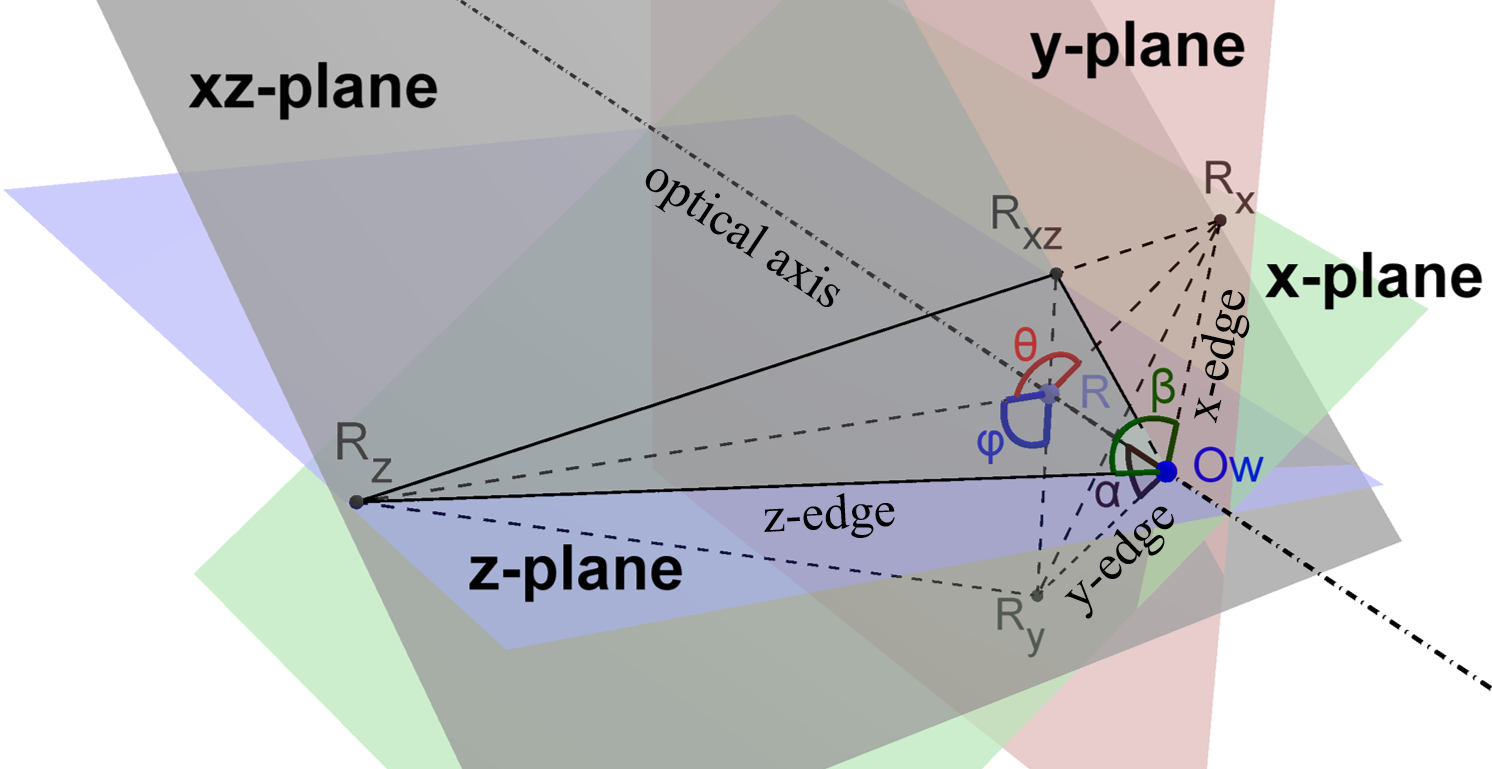}
  \caption{The Trove structure in space when all edges point away from the focal point}
  \label{3AxisProjection1}
\end{figure}

Three possible cases exist: all edges of the imaginary box point away from the focal point; only one edge of the imaginary box is perpendicular to optical axis; only one edge of the imaginary box points towards the focal point. One of such configurations is depicted in Fig.~\ref{3AxisProjection1}. The vertical edge of the imaginary box intersects the optical axis with an angle $\alpha$. The orientation of the imaginary box can be recovered by finding the angle $\alpha$. The following contents discuss the approaches to find $\alpha$ in the three configurations.

\subsection{All Edges Point Away from Focal Point}\label{subsecAllEdgeAway}

When all edges point away from the focal point, $ac>0$. An illustration of this case is shown in Fig.~\ref{3AxisProjection1}. Let the green plane be the x-plane, the red plane be the y-plane and the blue plane be the z-plane. The long dashed line is the optical axis of the camera. Those three planes are determined by the focal point and the projections of the respective edges. Since the projection of the vertex is on the principal point, the x-, y- and z-plane have a mutual intersection which is the optical axis. The vertex of the object must also be on the optical axis. Denote the vertex of the imaginary parallelepiped in the object frame as $O_w$. Suppose $O_{w}R_{y}$ is on the y-edge of the imaginary parallelepiped, where $R_{y}$ is on the y-plane. Construct xz-plane that passes through $O_w$ and are perpendicular to $O_{w}R_{y}$. $O_{w}R_{y}$ has an angle of $\alpha$ with the optical axis.

As being pointed out in the previous section that the vertical edge must point away from the focal point, $\alpha$ is constrained in the range of $(0,\pi/2)$. Construct a perpendicular line to the optical axis from $R_{y}$ intersecting the optical axis at $R$. The xz-plane would intersect the x- and y-plane respectively at two lines: $O_{w}R_{x}$ and $O_{w}R_{z}$. $R_x$ and $R_z$ are chosen so that $R_{x}R$ and $R_{z}R$ are perpendicular to the optical axis, hence $R, R_x, R_y$ and $R_z$ are coplanar. Denote the angle $\angle R_y R R_z$ as $\varphi$ and $\angle R_z R R_x$ as $\theta$.

Since the x- and z-edge are perpendicular to the y-edge, they must be on the xz-plane. Further, the x-edge must be on the intersection line of x-plane and xz-plane. Therefore, $O_{w}R_{x}$ must be collinear with the x-edge. Similarly, $O_{w}R_{z}$ must be collinear with the z-edge. In such a case, $\angle R_{x}O_wR_{z}$ should equal to the known angle $\beta$ between the two horizontal edges of the object. Now the problem becomes straightforward that to find an $\alpha$ given $\angle R_{x}O_wR_{z}=\beta$.

\begin{figure}[!htb]
  \centering
  \includegraphics[width = \linewidth]{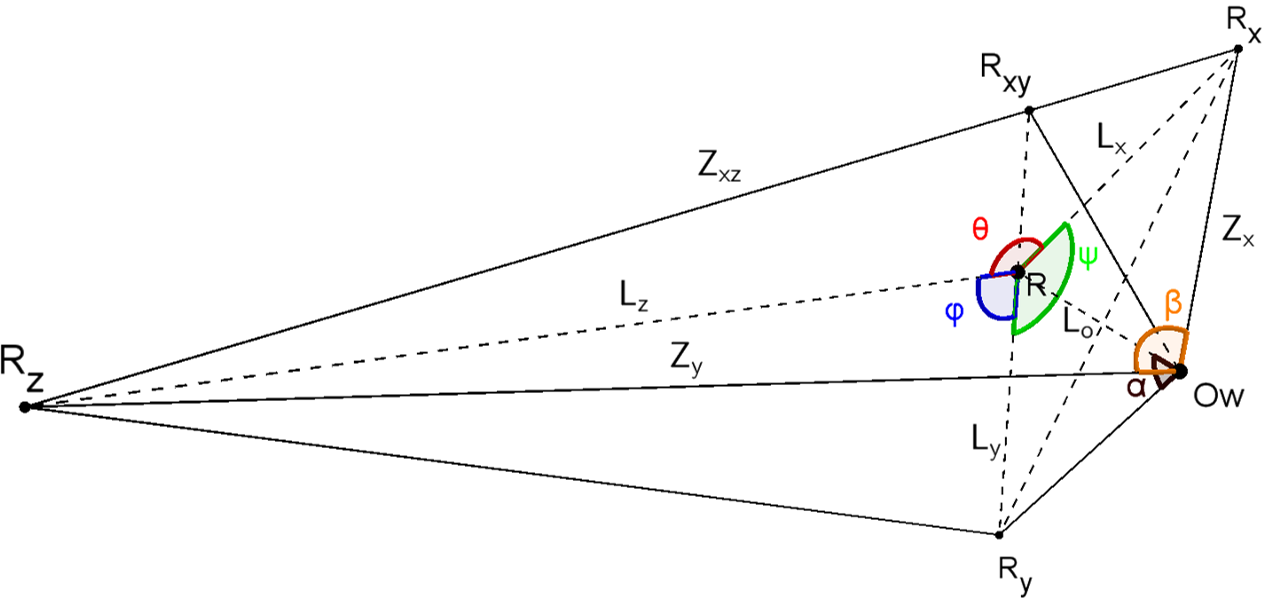}
  \caption{A closer inspection near the vertex}
  \label{3AxisProjection1Closer}
\end{figure}

A closer inspection of the space near $R$ is illustrated in Fig.~\ref{3AxisProjection1Closer}. Note that $R_{x}R$, $R_{y}R$ and $R_{z}R$ are perpendicular to $O_wR$. The angles of $\varphi$ is the included angles between y-plane and z-plane, and the angle of $\theta$ is the included angles between x-plane and z-plane. Since x-, y- and z-plane all intersect at the optical axis, $\varphi$ is actually the included angle between projected y-edge and z-edge in the image while $\theta$ is the included angle between projected x-edge and z-edge. Since the slope of edge projections has been obtained from edge detection, $\theta,\varphi$ and $\psi$ are known.

As denoted in Fig.~\ref{3AxisProjection1Closer}, let $|R_{x}O_w|$ be $Z_x$, $|R_{y}O_w|$ be $Z_y$, $|R_{x}R_y|$ be $Z_{xy}$, $|R_xR|$ be $L_x$, $|R_yR|$ be $L_y$, $|R_z R|$ be $L_z$ and $|O_wR|$ be $L_o$. Since plane $R_xR_zO_w$ is perpendicular to $R_yO_w$, one has:
\begin{align}
|R_xR_y|^2 = Z_x^2 + |R_yO_w|^2, |R_zR_y|^2 = Z_z^2 + |R_yO_w|^2. \label{RightTriangle1and2}
\end{align}
Note that plane $R_xR_zR_y$ is perpendicular to $RO_w$, one has:
\begin{align}
|R_yO_w| = L_o/\cos{\alpha},\,& |R_yR| = L_o\tan{\alpha}, \label{RightTriangle3and4}\\
Z_x^2 = L_o^2+L_x^2,\,& Z_z^2 = L_o^2+L_z^2. \label{RightTriangle5and6}
\end{align}
Applying the law of cosines for triangles to $\triangle R_xR_yR$, $\triangle R_zR_yR$, $\triangle R_xR_zO_w$ and $\triangle R_xR_zR$ gives:
\begin{align}
|R_xR_y|^2 &= L_x^2 + L_y^2 - 2L_xL_y\cos\psi, \label{cosines1}\\
|R_zR_y|^2 & = L_z^2 + L_y^2 - 2L_zL_y\cos{\varphi}, \label{cosines2}\\
Z_{xz}^2 & = Z_x^2 + Z_z^2 - 2Z_xZ_z\cos{\beta}, \label{cosines3}\\
Z_{xz}^2 & = L_x^2 + L_z^2 - 2L_xL_z\cos{\theta}. \label{cosines4}
\end{align}
By substituting \eqref{RightTriangle1and2}, \eqref{RightTriangle3and4}, and \eqref{RightTriangle5and6} into \eqref{cosines1} and \eqref{cosines2}, it can be obtained that:
\begin{align}
L_o^2+\Big(\frac{L_o}{\cos{\alpha}}\Big)^2=&L_o^2\tan^2\alpha-2L_xL_o\tan{\alpha}\cos{\psi}, \label{Substituted1} \\
L_o^2+\Big(\frac{L_o}{\cos{\alpha}}\Big)^2=&L_o^2\tan^2\alpha-2L_zL_o\tan{\alpha}\cos{\varphi}. \label{Substituted2}
\end{align}
Combining \eqref{cosines3}, \eqref{cosines4}, \eqref{Substituted1} and \eqref{Substituted2}, one can derive:
\begin{equation}\label{solutionCase1}
\cos{\beta} = \dfrac{ 1+\dfrac{\cos{\theta}}{\tan^2{\alpha}\cos{\varphi}\cos{\psi}} }{\sqrt{\Big(1+\dfrac{1}{\tan^2{\alpha}\cos^2{\varphi}}\Big)\Big(1+\dfrac{1}{\tan^2{\alpha}\cos^2\psi}\Big)}}.
\end{equation}
By denoting $\tan{\alpha}$ as $x$, $\cos{\beta}$ as $n$, $\cos{\theta}$ as $m$, $\cos{\varphi}$ as $p$ and $\cos{\psi}$ as $q$, \eqref{solutionCase1} can be rewritten as:
\begin{equation}\label{transformedSolution}
  \frac{m^2-n^2}{p^2q^2}\frac{1}{x^4}+\Big(\frac{2m}{pq}-\frac{n^2}{p^2}-\frac{n^2}{q^2}\Big)\frac{1}{x^2}+1-n^2=0,
\end{equation}
the solution of which must satisfy:
\begin{equation}\label{eq:solutionCriteria1}
  n\Big(1 + \frac{m}{x^2pq}\Big) > 0 \text{ or } n=1 + \frac{m}{x^2pq}=0. 
\end{equation}
Denote $(n^2p^2\!+n^2q^2\!-2mpq)^2\!-4(m^2\!-n^2)(1-n^2)p^2q^2$ as $\Delta$. Solving \eqref{transformedSolution} gives:
\begin{equation}\label{quadraticFormula}
\frac{1}{x^2}=\frac{(n^2p^2+n^2q^2-2mpq)\pm\sqrt{\Delta}}{2(m^2-n^2)}.
\end{equation}
Note that $x=\tan\alpha,\alpha\in(0,\pi/2)$. The solution uniqueness of $1/x^2$ indicates that of $\alpha$. 

It is obvious that the existence of a solution is guaranteed in that the image is from the projection of an existing instance. As having been proved in Section \ref{subsubsecFeatureProperties}, $n=\cos{\beta}>\cos\theta=m, pq>0$ in this case. Since $1/x^2$ is always greater than 0, only one feasible solution is available when $m^2-n^2<0$ (so that only one solution is greater than zero). When $m^2-n^2>=0$, both solutions are positive and $m<0$. Criteria \eqref{eq:solutionCriteria1} is then used to examine the feasibility of the solution. Denote the two solutions as $1/x^2_1$ and $1/x^2_2$. Study the product:
\begin{align}
  & \Big(1 + \frac{m}{x^2_1pq}\Big)\Big(1 + \frac{m}{x^2_2pq}\Big) \nonumber\\
= & \frac{1}{x^2_1}\frac{1}{x^2_2}\frac{m^2}{p^2q^2} + \frac{m}{pq}\Big(\frac{1}{x^2_1}+\frac{1}{x^2_2}) + 1 \nonumber\\
= & \frac{-m^2n^2-n^2+mn^2(p^2+q^2)/pq}{(m^2-n^2)}. \label{productX1X2Criteria}
\end{align}
When $n=0$, $\Delta=0$ and only one solution is available. When $n\neq0$ and $m^2-n^2>=0$, $m$ must be negative. Thus, product \eqref{productX1X2Criteria} must be smaller than zero and only one solution satisfies \eqref{eq:solutionCriteria1}. Therefore, only one feasible solution is available in the case where $m^2-n^2>=0$.

$\beta$ is known as priori knowledge and $\theta, \varphi$ are found by edge detection. In summary, $\alpha$ is uniquely determined in a closed form as:
\begin{equation}\label{alphaSolution1}
\alpha = \begin{cases}
           \arctan\Big(\sqrt{ \frac{ 2(m^2-n^2) }{ (n^2p^2+n^2q^2-2mpq)-\sqrt{\Delta} } }\Big), & \mbox{for } n>0 \\
           \arctan\Big(\sqrt{ \frac{-m}{pq} }\Big), & \mbox{for } n=0 \\
           \arctan\Big(\sqrt{ \frac{ 2(m^2-n^2) }{ (n^2p^2+n^2q^2-2mpq)+\sqrt{\Delta} } }\Big), & \mbox{for } n<0.
         \end{cases}
\end{equation}

\subsection{One Edge Perpendicular to Optical Axis}\label{subsecOneEdgePer}
\begin{figure}[!htb]
  \centering
  \includegraphics[width = \linewidth]{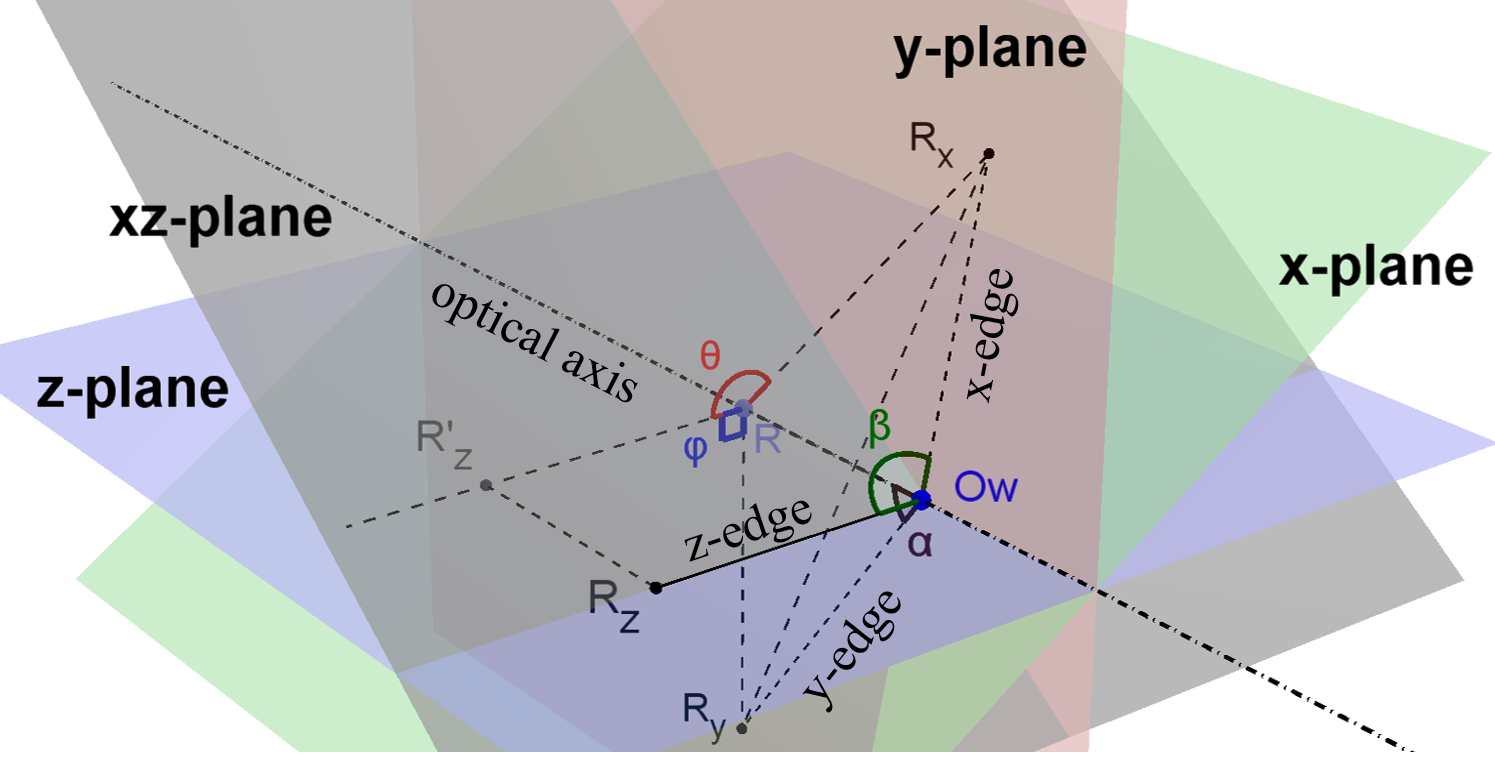}
  \caption{The Trove structure in space with one edge perpendicular to optical axis}
  \label{3AxisProjection2}
\end{figure}

Without loss of generality, suppose z-edge is perpendicular to the optical axis. An illustration of this case is shown in Fig.~\ref{3AxisProjection2}. As having been mentioned in the earlier Section \ref{subsubsecFeatureProperties}, the only possible range for $\beta$ in this case is $(\pi/2,\pi)$ and $\varphi$ must be $\pi/2$. The problem becomes much simpler with a straightforward unique solution:
\begin{equation}\label{alphaSolution2}
\alpha=\arccos(\tan\theta/\tan\beta),
\end{equation}
which is true for either the x-edge or z-edge is perpendicular to the optical axis. Because $n$ is always negative in this case, \eqref{alphaSolution2} gives the same result as \eqref{alphaSolution1}.

\subsection{One Edge Points to Focal Point}\label{subsecOneEdgeTo}
\begin{figure}[!htb]
  \centering
  \includegraphics[width = \linewidth]{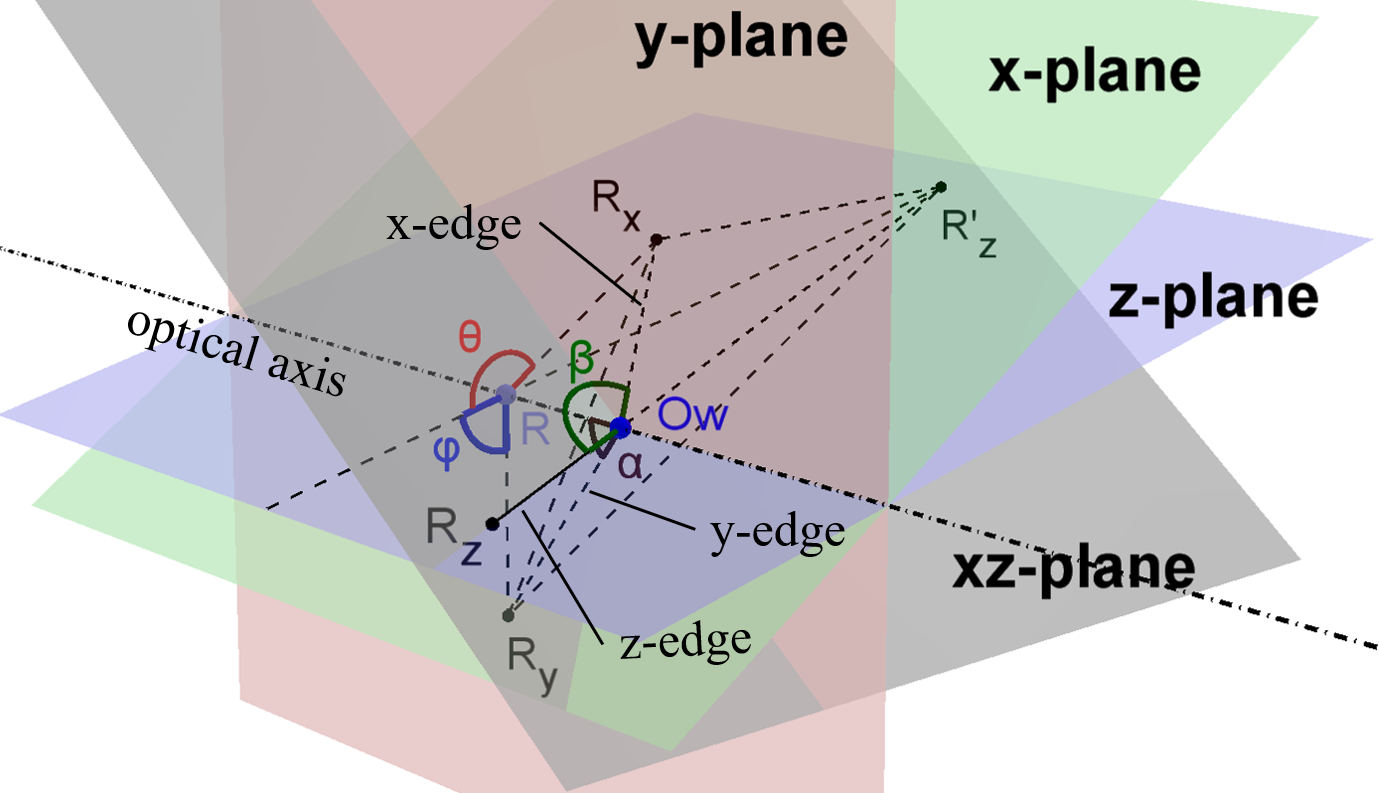}
  \caption{The Trove structure in space when one edge points to the focal point}
  \label{3AxisProjection3}
\end{figure}

As having been mentioned in the earlier Section \ref{subsubsecFeatureProperties}, the only possible range for $\beta$ in this case is $(\pi/2,\pi)$. It has properties that $\beta>\theta$ and $pq<0$. Without loss of generality, suppose the z-edge $O_wR_z$ points towards the focal point. An illustration of this configuration is shown in Fig.~\ref{3AxisProjection3}. $O_wR_z$ is extended in the opposite direction to $R'_z$ so that $R'_zR\perp RO_w$. All the other notations share the same definition as Section \ref{subsecAllEdgeAway}. Following a similar method, one can obtain:
\begin{equation}\label{solutionCase3}
\cos{\beta} = -\dfrac{ 1\!+\!\dfrac{\cos{\theta}}{\tan^2{\alpha}\cos{\varphi}\cos{\psi}} }{\sqrt{\Big(1\!+\!\dfrac{1}{\tan^2{\alpha}\cos^2{\varphi}}\Big)\Big(1\!+\!\dfrac{1}{\tan^2{\alpha}\cos^2\psi}\Big)}},
\end{equation}
the solution of which must satisfy:
\begin{equation}\label{eq:solutionCriteria2}
  1 + \frac{m}{x^2pq} > 0.
\end{equation}
This inequality is easily satisfied since $m<0, pq<0$. Note that the only difference from \eqref{solutionCase3} to \eqref{solutionCase1} is a minus sign. In solving $\alpha$, the equation is exactly the same as \eqref{quadraticFormula}. Since $n^2$ is always smaller than $m^2$ when $n<0$, $\alpha$ will have two feasible solutions in a closed form as:
\begin{equation}\label{alphaSolution3}
\alpha_{1,2}=\arctan\Big(\sqrt{ \frac{ 2(m^2-n^2) }{ (n^2p^2+n^2q^2-2mpq)\pm\sqrt{\Delta} } }\Big).
\end{equation}
Two roots will be different when:
\begin{equation}\label{deltaZero}
\Delta\neq0\Rightarrow\tan^2\beta = -\frac{(\tan\varphi+\tan\psi)^2}{4\tan\varphi\tan\psi}. 
\end{equation}

In summary, $\alpha$ can be uniquely determined in a closed form in all cases except the case where $\beta\in(\pi/2,\pi),\cos\varphi\cos\psi<0$ and \eqref{deltaZero} all hold. This exceptional case will be discussed at the end of the following subsection.

\subsection{Recover Attitude from $\beta$}

\begin{figure*}
  \centering
  \includegraphics[width = \linewidth]{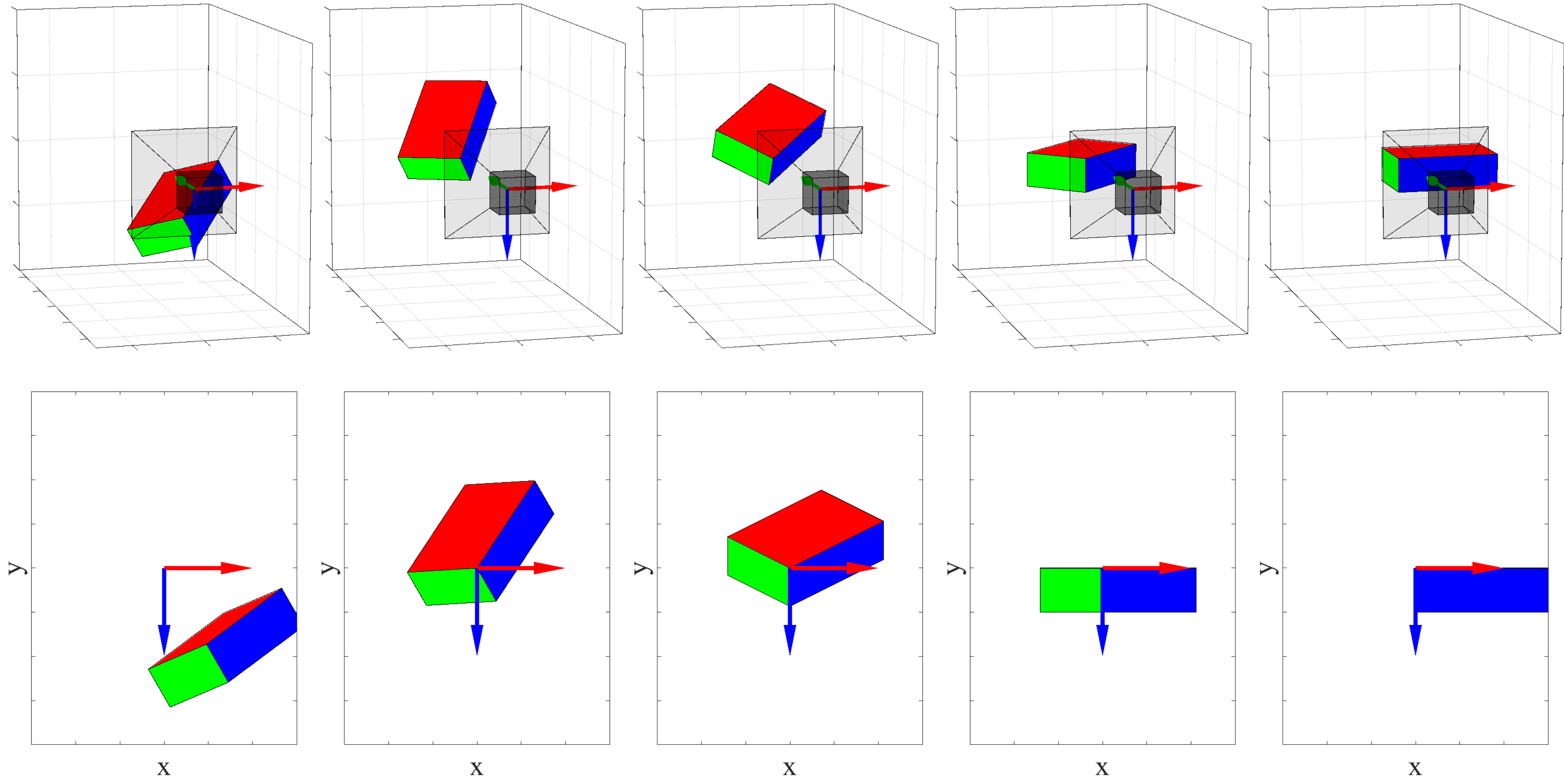}
  \caption{Rotate an object to align the object frame with the camera frame. The top row displays the camera and the object in 3D space. The bottom row displays the projected object in the image. Each column represents a phase during rotation in a sequence from left to right. }
  \label{objProjSeq}
\end{figure*}

A rotation matrix $\bm{R}_\textnormal{a}$ is defined in the camera frame that aligns the object frame with the camera frame. Note that $\bm{R}_\textnormal{a}$ is identical to the attitude of the camera in the object frame. Fig. \ref{objProjSeq} illustrates a sequence that rotates an object into alignment with the camera frame. The object is represented by a cuboid and the camera is represented by a horizontal pyramid. The top row displays the camera and the object in 3D space. The bottom row displays the projected object in the image. The origin of the camera frame is at the focal point of the camera. The red, blue and green arrows represent the x-, y- and z-axis of the camera frame, respectively. Each column represents a phase during rotation in a sequence from left to right. First column: the original positions of the object and the camera are depicted; second column: rotation, denoted as a matrix $\bm{R}_\gamma$, is applied around the focal point so that the vertex is projected onto the principal point; third column: rotation, denoted as a matrix $\bm{R}_\varphi$, is applied so that y-edge is projected vertical in the image; third column: rotation, denoted as a matrix $\bm{R}_\alpha$, is applied so that y-edge is aligned with the y-axis; forth column: rotation, denoted as a matrix $\bm{R}_\beta$, is applied so that x-edge is aligned with the x-axis. Thus far, the object frame is in alignment with the camera frame. The following paragraph elaborates on finding $\bm{R}_\gamma,\bm{R}_\varphi,\bm{R}_\alpha,\bm{R}_\beta$ and subsequently the attitude of the camera.

As having been previously derived in Section \ref{subsectionStandardModel}, the rotation axis and angle of $\bm{R}_\gamma$ can be determined by \eqref{gammaSolu}. Rotation $\bm{R}_\varphi$ is applied so that the y-edge is projected vertical in the image. The rotation axis is the z-axis of the camera frame while the rotation angle is the included angle by the projected y-edge and the y-axis of the image frame. The rotation axis of $\bm{R}_\alpha$ is the x-axis and the angle is $(\alpha-\pi/2)$. By using the notations in Fig. \ref{3AxisProjection1}, rotation $\bm{R}_\beta$ is to align the x-edge with the x-axis of the camera frame after the xz-plane is rotated horizontal. Denote the rotation angle as $\beta'$. Obviously, $\beta'=\pi/2-\angle R_{xy}O_wR_x$. It can easily be calculated that for all aforementioned three cases in Section \ref{subsecAllEdgeAway}, \ref{subsecOneEdgePer} and \ref{subsecOneEdgeTo} $\beta'=\arctan(-\cos\alpha/\tan\psi)$. Thus, $\bm{R}_\textnormal{a}$ can be expressed as:
\begin{equation}\label{camAttSolu}
  \bm{R}_\textnormal{a}=\bm{R}_\beta\bm{R}_\alpha\bm{R}_\varphi\bm{R}_\gamma.
\end{equation}

There is still one more process before the actual attitude of a camera is recovered. Whenever a TROVE feature is detected, two interpretations are possible as illustrated in Fig. \ref{FeatureInterpretation}(b) and \ref{FeatureInterpretation}(c). If the attitude in Fig. \ref{FeatureInterpretation}(b) is expressed as \eqref{camAttSolu}, the attitude in Fig. \ref{FeatureInterpretation}(c) can be expressed as $\bm{R}_a=\bm{R}_\beta\bm{R}_\alpha\bm{R}_\varphi\bm{R}_{z,\pi}\bm{R}_\gamma$, where $\bm{R}_{z,\pi}$ is the rotation around the z-axis by $\pi$. Two methods can be applied to discard the incorrect interpretation.

Method 1: if an image captures more than one TROVE feature that has parallel frames to each other or the same feature is captured by two parallel cameras simultaneously (the case of a stereo camera), the result from the correct interpretation will always be consistent but the result from the incorrect one will vary. One simply accepts the consistent result as the estimation. Proof is give as follows. If two features have parallel frames, the attitude estimate should be the same. Let features 1 and 2 yield the same correct attitude estimate:
\begin{align}
  \bm{R}_a&=\bm{R}_{\beta1}\bm{R}_{\alpha1}\bm{R}_{\varphi1}\bm{R}_{\gamma1},\label{camAttTwoCorr1}\\
  \bm{R}_a&=\bm{R}_{\beta2}\bm{R}_{\alpha2}\bm{R}_{\varphi2}\bm{R}_{\gamma2},\label{camAttTwoCorr2}
\end{align}
where the subscripts 1 and 2 represents the two features. Then the incorrect estimate can be expressed as:
\begin{align}
  \bm{R}_{a1}&=\bm{R}_a\bm{R}^\intercal_{\gamma1}\bm{R}_{z,\pi}\bm{R}_{\gamma1},\label{camAttTwoIncorr1}\\
  \bm{R}_{a2}&=\bm{R}_a\bm{R}^\intercal_{\gamma2}\bm{R}_{z,\pi}\bm{R}_{\gamma2}.\label{camAttTwoIncorr2}
\end{align}
Suppose the incorrect estimates are equal. Equating \eqref{camAttTwoIncorr1} $=$ \eqref{camAttTwoIncorr2} yields:
\begin{equation}\label{camAttTwoIncorrEqua}
  \bm{R}_{z,\pi} = \bm{R}_{\gamma1}\bm{R}^\intercal_{\gamma2}\bm{R}_{z,\pi}\bm{R}_{\gamma2}\bm{R}^\intercal_{\gamma1}.
\end{equation}
It can be proved that the only possible instances are $\bm{R}_{\gamma1}=\bm{R}_{\gamma2}$ or $\bm{R}_{\gamma2}\bm{R}^\intercal_{\gamma1}$ is a rotation around the z-axis as $\bm{R}_{z,\pi}$. Recall that $\bm{R}_{\gamma}$ is determined by the position of the vertex in the image. Different features cannot share the same vertex in one image and the vertex of a structure will be projected differently in the images by different parallel cameras. Therefore $\bm{R}_{\gamma1}\neq\bm{R}_{\gamma2}$. $\bm{R}_{\gamma}$ is a rotation around the axis $(x,y,0)$. It can also be proved that $\bm{R}_{\gamma2}\bm{R}^\intercal_{\gamma1}$ cannot be a rotation around the z-axis unless $\bm{R}_{\gamma1}=\bm{R}_{\gamma2}$. Hence, the assumption must be false and the incorrect estimates will always be different across different features in an image or across the same features in different images captured by parallel cameras.

Method 2: the interpretation of Fig. \ref{FeatureInterpretation}(b) is correct only if the camera has a negative pitch angle, namely the z-axis of the camera frame points downward. Similarly, the interpretation of Fig. \ref{FeatureInterpretation}(c) is correct only if the camera has a positive pitch angle. If the inclination of the camera to the horizontal surface is known, one can directly discard the incorrect interpretation. Method 2 appears to have a paradox where one needs to know the attitude to estimate the attitude. By the help of priori estimates or other sensors such as accelerometers, one can distinguish whether the camera points upward to downward. Even a rough estimate can help discard the incorrect interpretation for the interpretations have obvious discrepancies on the sign of the pitch angle.

Another rare but possible case where the solution is not unique is that $\beta\in(\pi/2,\pi),\cos\varphi\cos\psi<0$ and \eqref{deltaZero} all hold as described in Section \ref{subsecOneEdgeTo}. That means the angle formed by the two horizontal edges are obtuse and one of the horizontal edge points towards the observer. When one edge points towards the observer, one face will usually be occluded. But it is still possible that the three concerning faces are visible. In this case, $\alpha$ has two solutions unless \eqref{deltaZero} does not hold. One characteristic of this case is that one and only one of $\varphi$ and $\psi$ is an acute angle. One only needs to take measures when this characteristic has been observed. The two methods to discard the incorrect solutions are almost the same as the previous ones to discard incorrect interpretations. The only difference is in Method 2 that the pitch angle could be of the same sign in this case, therefore a relatively more accurate priori estimate is needed.

\subsection{Recover $\beta$ from Attitude}

If one knows the relative attitude to the imaginary box, recovering $\beta$ is straightforward as it is uniquely determined by $\alpha$. Suppose that the relative attitude $\bm{R}_a$ is obtained. Since $\bm{R}_\gamma,\bm{R}_\varphi$ can be directly estimated from the image, one can easily obtain from \eqref{camAttSolu} that:
\begin{equation}\label{betaSolu}
  \bm{R}_\beta\bm{R}_\alpha=\bm{R}_a\bm{R}^\intercal_\gamma\bm{R}^\intercal_\varphi.
\end{equation}
Expanding the rotation matrices yields:
\begin{equation}
\bm{R}_\beta\bm{R}_\alpha =
  \begin{bmatrix*}[c]
    \cos\beta'     & -\sin\beta'\cos\alpha        & -\sin\beta'\sin\alpha    \\
    0               & -\sin\alpha                   & \cos\alpha                \\
    -\sin\beta'    & -\cos\beta'\cos\alpha        & -\cos\beta'\sin\alpha
  \end{bmatrix*}.
\end{equation}
With the value of $\bm{R}_\beta\bm{R}_\alpha$ available, $\alpha$ can be directly obtained.

The angles of $\theta, \varphi$ and $\psi$ are obtained from image processing. When $\varphi,\psi\in(\pi/2,\pi)$, $\beta$ can be computed by \eqref{solutionCase1}. When one of $\varphi$ and $\psi$ is $\pi/2$, $\beta$ can be computed by \eqref{alphaSolution2} as $\beta=\arccos(\cos\theta/\cos\alpha)$. When one of $\varphi,\psi$ is less than $\pi/2$, $\beta$ can be computed by \eqref{solutionCase3}.

\subsection{Recover Position}

The vertex of a TROVE structure is often a stationary point of the environment, such as the corner of a room and the apex of a building. For a robot to autonomously navigate through a certain environment, the relative position to the environment is often much more important than the absolute position in the Earth frame.

The camera is a binocular camera. By the disparity of the same vertex in two images, the relative position of the camera to a stationary point in space is obtained. The position of a vertex in each image is not detected by investigating the pixel itself or neighbouring groups like other stereo matching methods. The vertex is determined as the intersection of three detected lines, each of which is estimated by hundreds of pixels. Therefore, the accuracy can easily reach a sub-pixel level. Despite errors of installation and calibration, the result still remains in high accuracy, detail of which is given in Section \ref{SecExperiment}.

\section{Experiment and Result}\label{SecExperiment}

In this section, the experiments that evaluate the accuracy and effectiveness of the proposed method are presented. Attitude estimation is verified by comparison with the ground truth and the conventional method of the Complementary Filter. Position estimation is verified by direct comparison with the ground truth. The camera and IMUs are mounted on a board. The board is manually moved around to simulate a robot navigating in an unknown environment. Infra-red sensitive markers are also mounted on the board for the motion tracking system (OptiTrack) to capture poses which are then used the ground truth. A colored cuboid is placed horizontally on the ground as the object the robot refers to for ego-states estimation. All the data are recorded and processed online. The details of the system architecture and experiment setup are enclosed in the following subsections. Since attitudes are measured as rotation from an initial pose, having a consistent initial pose in all sensors' frames is pivotal to evaluate the accuracy of estimates. The description of calibration process is also enclosed in the following subsection. It has been observed that without calibration the errors would be more than threefold.

\subsection{Experiment Setup}

\begin{figure}[!htb]
  \centering
  \includegraphics[width = 2 in]{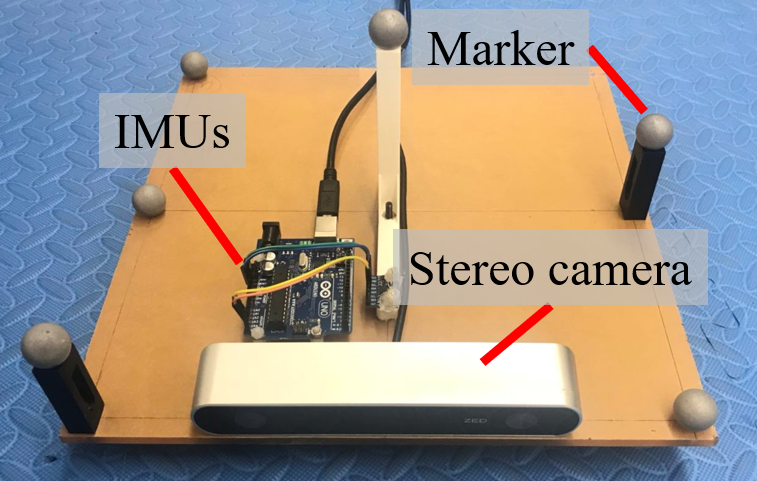}
  \caption{A board mounted with a binocular camera and IMUs. }
  \label{BoardWithZEDIMU}
\end{figure}

The saturation, white balance and sharpness of the camera have all been fixed so that the color in RGB values is consistent. The resolution of the camera is set as $1920\times1080$ and $1280\times720$, respectively. As shown in Fig.~\ref{BoardWithZEDIMU}, the camera together with IMUs and infrared sensitive markers are rigidly mounted onto a flat board. The top face of the board is referred as the horizontal plane of the board. The camera is a ZED Camera manufactured by Stereolab. The baseline of the stereo camera is $\SI{120}{\milli\meter}$. The focal length of the lens is 1049 and 702 in pixels at a resolution of $1920\times1080$ and $1280\times720$, respectively. The IMU sensor is MPU6050, which includes an accelerometer of a range of $\pm2$ g and a gyroscope of a range of $\pm\SI{2000}{\degree\per\second}$. The IMU sensor has also been calibrated to offset the misalignment of the horizontal plane between the IMU frame and the Earth frame. The infrared sensitive markers form a rigid body which is registered in the optical tracking software. The optical tracking software together with the infrared cameras are the OptiTrack system developed by NaturalPoint Inc. OptiTrack cameras can capture the markers and offers the 6-DoF states of the rigid body in real-time. The infrared cameras are mounted on the walls of the laboratory. The accuracy is within millimeters and the latency is at most $\SI{4.2}{\milli\second}$. The measurement from the OptiTrack system is used as the ground truth in this paper. In the experiment, the Earth frame actually refers to the frame that is defined in the OptiTrack system.

\begin{figure}[!htb]
  \centering
  \includegraphics[width = 2.5 in]{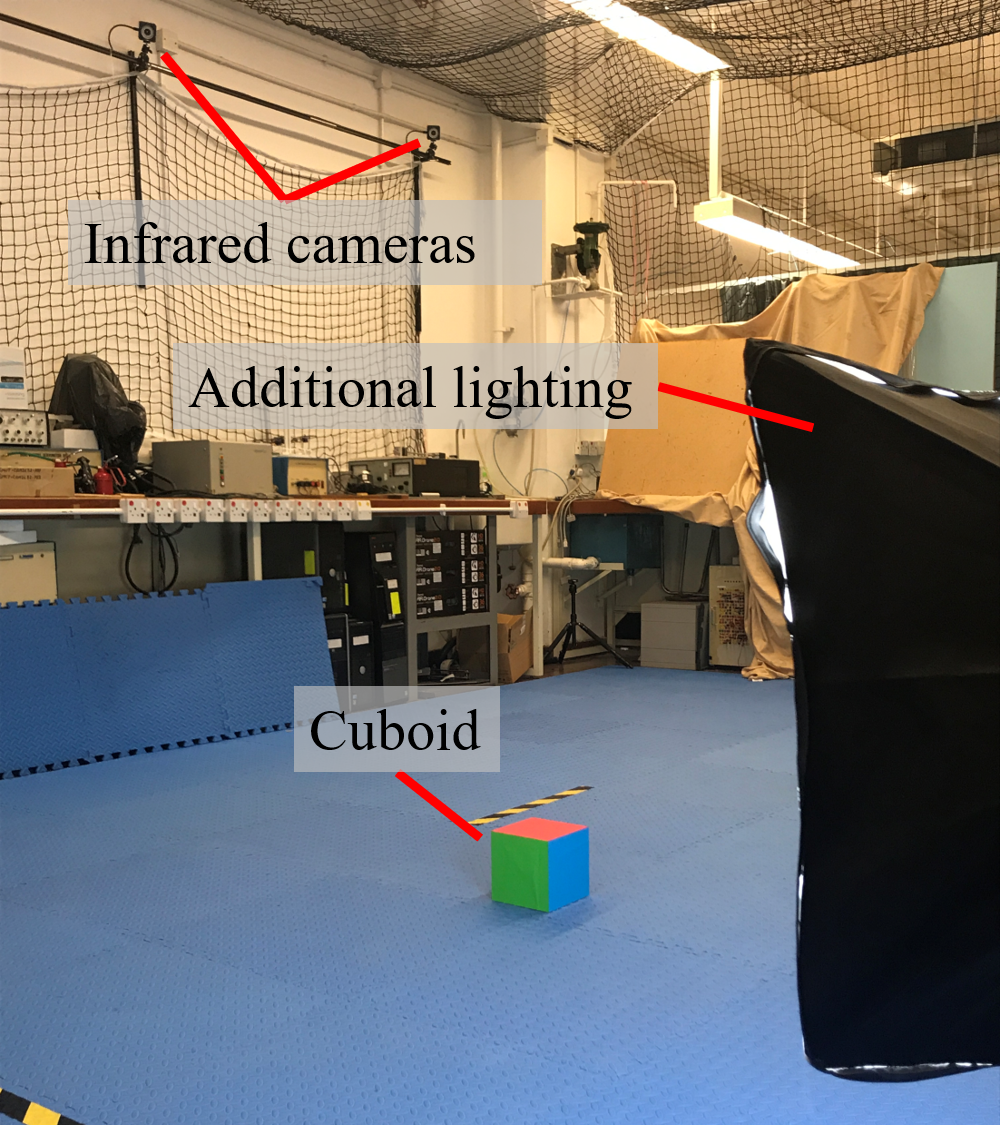}
  \caption{This figure shows the experiment setup in the laboratory with additional lighting at side. Infrared cameras of the tracking system are mounted on the laboratory walls. }
  \label{LabEnvironment}
\end{figure}

A cuboid is placed on the ground of the laboratory, as shown in Fig.~\ref{LabEnvironment}. Three adjacent faces of the cuboid are colored in red, green and blue, respectively. In the experiment, captured pixels of the red, green and blue face are around $[255,115,0]$, $[0,250,80]$ and $[0,100,215]$ in RGB values, respectively. As previously discussed, color segmentation is based on the thresholds of chromaticity and intensity. Those colors share similar characteristics so that the standards are the same for each color as $150$ out of $255$ in intensity and $\SI{51}{\percent}$ in chromaticity. Additional lighting is applied to the side of the cuboid to compensate the insufficient luminosity in the laboratory. It is apparent that all the edges of the cuboid passing through the concerning vertex are perpendicular to each other. As having been proved in Section \ref{subsubsecFeatureProperties}, the edges of the imaginary box must all point away from the focal point. With $\beta = \pi/2$, \eqref{alphaSolution1} yields:
\begin{equation}
\alpha = \arctan\Bigg(\sqrt{ \dfrac{-\cos{\theta}}{\cos{\varphi}\cos\psi} }\Bigg).
\end{equation}

\begin{figure}[!htb]
  \centering
  \includegraphics[width = \linewidth]{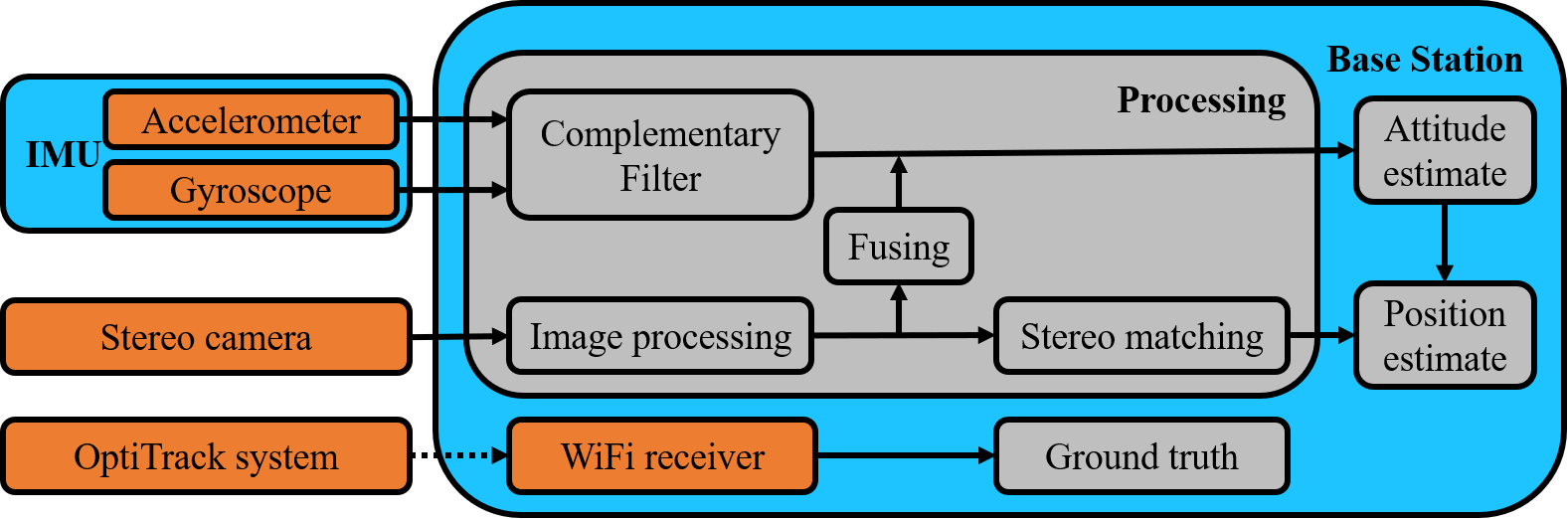}
  \caption{A diagram of system architecture. }
  \label{systemArchitecture}
\end{figure}

The data from the camera and IMUs are transmitted through a wire to a desktop computer as the base station. Then the data are processed and stored by the computer. The measurement from OptiTrack system is received through local wireless network by the same computer. Fig. \ref{systemArchitecture} illustrates the system architecture. Using the USB 3.0 interface of the stereo camera, images can be captured and transmitted at $\SI{60}{\hertz}$ at a resolution of $1280\times720$ and $\SI{30}{\hertz}$ at a resolution of $1920\times1080$. Both data from IMUs and OptiTrack system are received at $\SI{240}{\hertz}$. Whenever data are received, time stamps are recorded for synchronization. When IMU data are available, readings from the accelerometer and gyroscope are fused by the Complementary Filter for an attitude estimate. Whenever an attitude estimate from an image is available, the priori estimate is incremented by rotational rates and then fused with the image estimate. In the mean time, the disparity of the vertex on stereo images is found for position estimation. Attitude estimates are also used to obtain the position in the object frame.

The computer is of a moderate configuration with an Intel i7-7700 quad-core CPU running at $\SI{3.60}{\giga\hertz}$. No graphics card is used for the processing. The program is in C++ language utilizing multithreading and OpenCV library. Processing time (from Steps 2 to 23 in Alg. \ref{GeneralAlgo} and attitude estimation) is recorded for each image. For images at a resolution of $1920\times1080$, 90 pairs of stereo images are processed 20 times to take the average. For each pair of stereo images, namely two $1920\times1080$ images, it takes $\SI{6.534}{\milli\second}$ in total or $\SI{3.267}{\milli\second}$ for each image. For the resolution of $1280\times720$, 1000 pairs of stereo images are processed 20 times to take the average. For each pair, namely two $1280\times720$ images, it takes $\SI{2.823}{\milli\second}$ in total or $\SI{1.411}{\milli\second}$ for each image. Table \ref{AlgoTimer} shows the detailed time spent in the processing for each image. Compared with image processing, the other processing time is trivial. However, the image transmission time turns out to be much longer than the processing time. The update frequency is largely determined by a camera's capability of capturing and transmitting images. The latency of each image is about $\SI{25}{\milli\second}$ for both resolutions.

\begin{table}[ht]
\renewcommand{\arraystretch}{1.3}
\caption{Processing time for each image}
\label{AlgoTimer}
\centering
\begin{tabular}{lcc}
\hline\hline
\multirow{2}{*}{Process}    & \multicolumn{2}{c}{Time ($\SI{}{\milli\second}$)}\\
\hhline{~-|-}
                        & $1920\times1080$ & $1280\times720$ \\
\hline
Color Segmentation      & 1.360 & 0.638 \\
Edge Detection          & 1.826 & 0.679 \\
RANSAC Line Detection   & 0.032 & 0.044 \\
Attitude Estimation     & 0.050 & 0.050 \\
\hline
Total                   & 3.267 & 1.411 \\
\hline\hline
\end{tabular}
\end{table}

\subsection{Camera Calibration}

As discussed in \cite{weng1992camera}, camera distortion includes radial distortion, centering distortion, thin prism distortion and total distortion. In this paper only the edge pixels are of the concern, therefore the rectification is only applied to the found edge pixels just before line detection. Many tools can be utilized to calibrate distortion, one of which is the calibration tool provided by MATLAB as discussed in \cite{corke2011robotics}. The images (actually the coordinates of pixels) used by the proposed algorithm in this paper have all been rectified. Another calibration to be conducted is the attitude difference between the two lenses of the stereo camera, because the two lenses are not installed perfectly facing the same direction.

The last calibration is to ensure that the OptiTrack system frame, experiment board, the IMU and the camera share the same horizontal plane. To evaluate attitude estimate, inclinations to the horizontal plane are used. The reasons for referring to inclinations are: 1) the horizontal plane of all the concerning frames is verified while aligning the other axes of all frames can be challenging. Direct comparison of attitudes (such as comparison in Euler angles) will inevitably contain those systematic errors; 2) the attitude obtained from the conventional Complementary Filter cannot recover the yaw angle in the Earth frame. Thus comparing the attitude is not convincing; 3) that such information is pivotal to stably and accurately control a UAV. The attitude of an object is defined as the rotation from an initial orientation to the current one. In the experiment, the initial orientation of the camera is defined when the board's top face is placed horizontally, verified by a spirit level. Since the top face of the cuboid is placed horizontally which is also verified by a spirit level, the initial orientation of the board in the OptiTrack frame and the camera in the camera frame share the same horizontal plane. However, the top face of the board and the lenses are not horizontally parallel due to installation errors. Such a difference should be estimated and offset. In the experiment, cameras are placed at various positions and attitudes to capture images of the cuboid. Estimates from images are compared with the ground truth. The result shows that when in a resolution of $1920\times1080$ the attitude of lens with respect to the board is $(\SI{-2.44}{\degree}, \SI{1.19}{\degree}, \SI{-0.01}{\degree})$ in Euler angles in a rotation sequence of x-, y- and z-axis. When in a resolution of $1280\times720$ the difference is $(\SI{-2.39}{\degree}, \SI{1.17}{\degree}, \SI{0.03}{\degree})$. Without calibration the difference will introduce a bias of more than $\SI{2.6}{\degree}$ in inclination estimation.

\subsection{Fusing Method}

An IMU is one of the common devices to obtain attitudes for various types of robots. In the experiment, the attitude estimates from cameras are fused with those from IMUs to improve accuracy. Attitude estimates from IMUs are obtained by the Complementary Filter as discussed in \cite{mahony2005complementary}. A weighting $w_a$ is assigned to the measurements from the accelerometer in the Complementary Filter. In theory, the yaw angle, i.e. the rotation of an object around the vertical axis, cannot be determined solely by IMUs. But it can be estimated by the proposed method in the object frame.

Having estimates from cameras, one can fuse them for more accurate results. Even the estimates from cameras are already more accurate than IMUs. By such fusion the estimates from cameras are not only further improved in accuracy but also increased to $\SI{240}{\hertz}$. IMUs can run up to a thousand hertz while cameras can often only run at a few dozens of hertz. The fusion happens only when an image estimate is available. Suppose an image estimate, denoted as $\prescript{}{\textnormal{i}}{\hat{\bm{q}}}_k$ in a quaternion form, is available at time $k$. The previous estimate is $\hat{\bm{q}}_{k-1}$. By integrating the last rotational rates from the gyroscope during the period from $k-1$ to $k$, an increment in attitude can be obtained as $\prescript{}{\Delta}{\hat{\bm{q}}}_k$. An estimate based on rotational rates integration can be obtained as $\prescript{}{\textnormal{g}}{\hat{\bm{q}}}_k = \prescript{}{\Delta}{\hat{\bm{q}}}_k\hat{\bm{q}}_{k-1}$. The fusion is achieved by a Slerp interpolation \cite{dam1998quaternions} with a weighting $w_i$ assigned to image estimates:
\begin{equation}
  \hat{\bm{q}}_{k} = \prescript{}{\textnormal{i}}{\hat{\bm{q}}}_k(\prescript{}{\textnormal{i}}{\hat{\bm{q}}}^*_k\prescript{}{\textnormal{g}}{\hat{\bm{q}}}_k)^{1-w_i} = \prescript{}{\textnormal{i}}{\hat{\bm{q}}}_k(\prescript{}{\textnormal{i}}{\hat{\bm{q}}}^*_k\prescript{}{\Delta}{\hat{\bm{q}}}_k\hat{\bm{q}}_{k-1})^{1-w_i}\label{imageFuse},\\
\end{equation}
where $\bm{q}^*$ represents the conjugate of $\bm{q}$.

\subsection{Optimal Weighting}

\begin{figure*}
     \centering
     \begin{subfigure}[b]{0.5\linewidth}
         \centering
         \includegraphics[width=\linewidth]{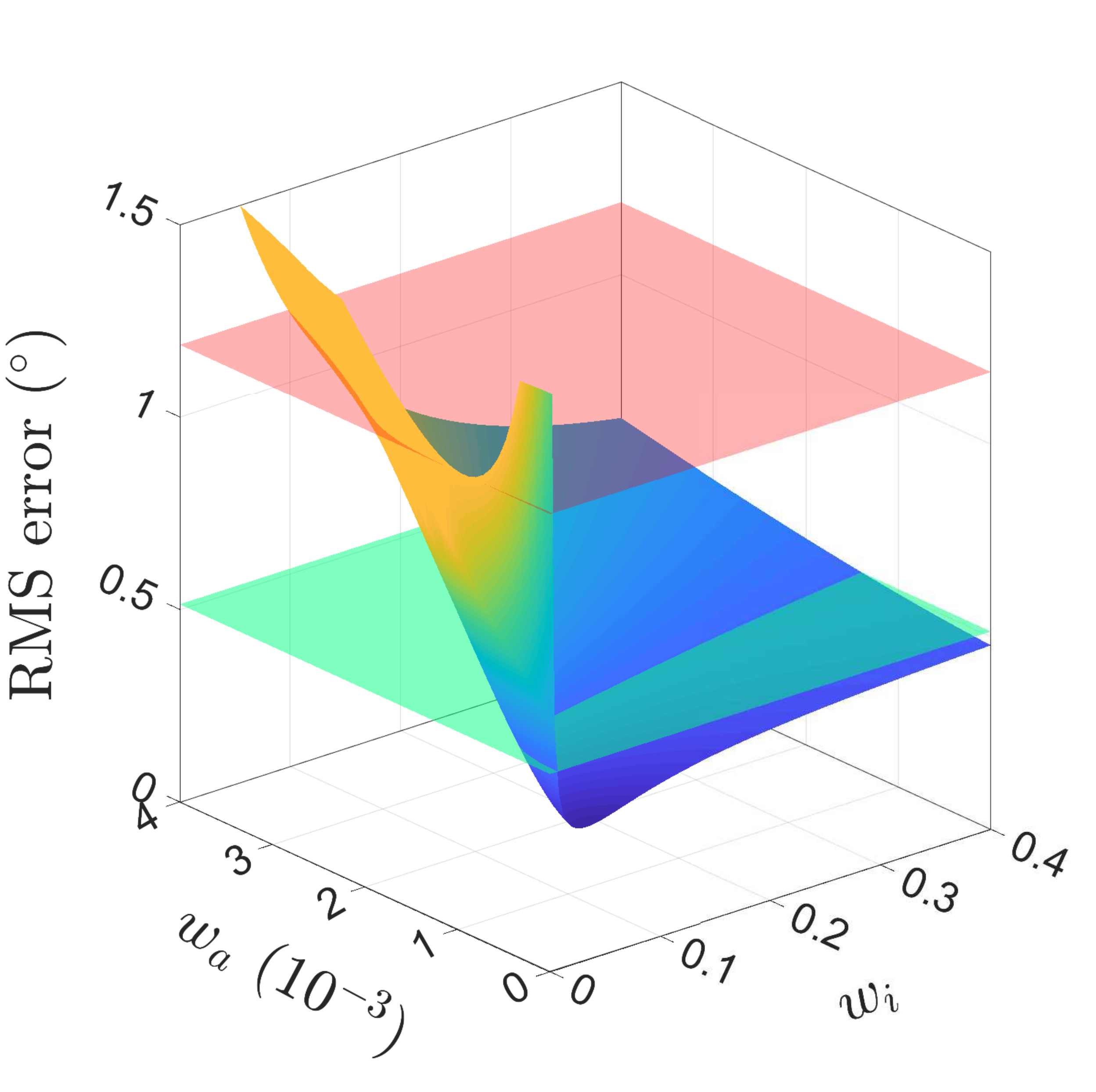}
         \caption{}
         \label{fuseWeight1080}
     \end{subfigure}%
     \begin{subfigure}[b]{0.5\linewidth}
         \centering
         \includegraphics[width=\linewidth]{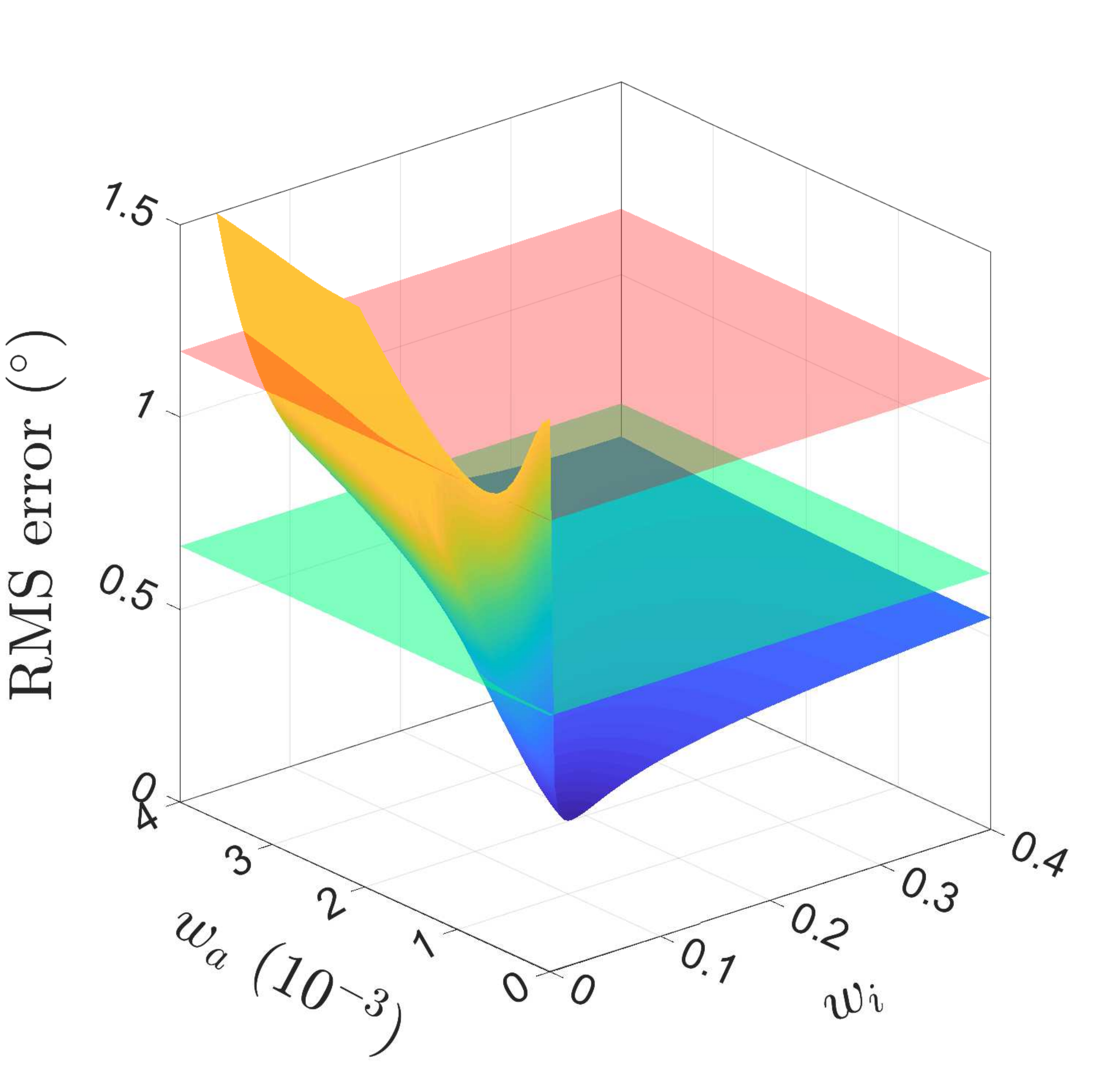}
         \caption{}
         \label{fuseWeight720}
     \end{subfigure}%
     \caption{Average error magnitude for various pairs of $w_i$ and $w_a$. (a) Images at a resolution of $1920\times1080$; (b) images at a resolution of $1280\times720$.}
     \label{fuseWeight}
\end{figure*}

To find the optimal weighting, pairs of $w_i\in[0,0.4]$ and $w_a\in[0,0.004]$ are applied to the fusion. Two sets of data are recorded in a period of $\SI{60}{\second}$, during which the board is constantly moved around. The data of $1920\times1080$ images are recorded at $\SI{30}{\hertz}$, consisting of 1801 frames of stereo images. The data of $1280\times720$ images are recorded at $\SI{60}{\hertz}$, consisting of 3601 frames of stereo images. IMU data are recorded at $\SI{240}{\hertz}$. The optical tracking data is referred to as the ground truth, and are recorded at $\SI{240}{\hertz}$.

Fig. \ref{fuseWeight}(a) and Fig. \ref{fuseWeight}(b) present the (Root-Mean-Square) RMS errors for various pairs of $w_i$ and $w_a$ by fusing images of resolutions of $1920\times1080$ and $1280\times720$, respectively. The z-axis is cut off at $\SI{1.5}{\degree}$. The red plane represents the error of the Complementary Filter with the optimal $w_a$ while the green plane represents the error of sole image analysis. It can be obviously observed from the figure that the results from the fusion adopting the optimal weighting is better than both the optimal Complementary Filter and the image analysis with regard to RMS errors. It is noteworthy that when $w_i=0$ the fusion method degrades to the Complementary Filter. The pair that results with the smallest error is selected for the latter experiment. For the Complementary Filter, the optimal weighting is $w_a=8\times10^{-4}$. For images of a resolution of $1920\times1080$, the optimal pair is $w_i=0.032$ and $w_a=0$. For images of a resolution of $1280\times720$, the optimal pair is $w_i=0.020$ and $w_a=0$

The result suggests that with image estimates the estimation will be most accurate without any contribution from the accelerometer. Not only more accurate estimates are obtained by the fusion, but also estimates' frequency increases up to $\SI{300}{\hertz}$. The fusion happens whenever an image is available. Therefore, the update interval is not constant after fusion. All the attitude estimation in the following experiments has adopted the optimal weighting.

\subsection{Attitude Estimation Result}

First, the camera is held by a tripod to keep a stationary pose when taking each image of the object. In this static scenario, image estimates are compared with the ground truth to verify the accuracy. Second, the board hovers and maneuvers simulating a navigating UAV to capture images dynamically. In this dynamic scenario, the image estimates suffer from motion blurs and latencies. Videos at $1920\times1080$ and $1280\times720$ are recorded. By comparing the estimates from the Complementary Filter, images, the fusion with the ground truth, the accuracy is evaluated. The rest of this section discusses the results in the two scenarios.

\subsubsection{Static Scenario}

\begin{table}
\renewcommand{\arraystretch}{1.3}
\caption{Inclination estimate errors ($^\circ$) of static images}
\label{statResult}
\centering
\begin{tabular}{lcccccc}
\hline\hline
  \multirow[l]{2}{*}{Resolution} &\multicolumn{3}{c}{Each image} & \multicolumn{3}{c}{Each pair}\\
  \cmidrule(lr){2-4} \cmidrule(lr){5-7}
  & $\overline{\lvert e\rvert}$& $\sigma$ & RMS & $\overline{\lvert e\rvert}$ & $\sigma$ & RMS\\
\hline
  $1920\times1080$ & 0.182 & 0.240 & 0.265& 0.161 & 0.193 & 0.222\\
\hline
  $1280\times720$ & 0.227 & 0.275 & 0.275 & 0.223 & 0.264 & 0.262\\
\hline\hline
\end{tabular}
\end{table}

Other than the data collected for calibration, another 145 pairs of stereo images at $1920\times1080$ are captured at various poses and distance to the observed object. Out of the total 290 images 278 have recognized the feature. For each image, the average of error magnitude $\overline{|e|}$, the standard deviation and the RMS error are all listed in Table \ref{statResult}. If one averages the estimates from the two images of each pair, obvious improvement can be achieved as shown in the table. Similarly, 89 pairs of stereo images at $1280\times720$ are captured and analyzed. Out of the total 178 images 173 have recognized the feature. As expected, the estimates are less accurate than images at $1920\times1080$ but the RMS errors are close between the two resolutions. After averaging two estimates for each pair, trivial improvement has been observed. The experiment indicates that the proposed method is of high accuracy, given that only a stereo camera is used. Images at $1920\times1080$ are more accurate than those at $1280\times720$. Averaging two estimates for each pair yields obvious improvement for images at $1920\times1080$, but trivial improvement for images at $1280\times720$. In the experiment it has been observed that below the accuracy level of $\SI{0.1}{\degree}$ the estimation is more susceptible to calibration and systematic errors.

\subsubsection{Dynamic Scenario}

In a more practical case, a navigating robot constantly changes its attitudes and positions. The moving of the camera and inconsistent latency of sensors all affect the result. The effectiveness and accuracy needs to be evaluated in such a dynamic scenario. The experiment also compares the accuracy between the conventional method and the proposed method.

For both the Complimentary Filter and the image fusion method, the initial attitude estimate takes the measurement from the OptiTrack system. By such means, the estimation spends much less time to converge at the beginning and it mitigates the influence of variations of initial conditions.
Because of such an adjustment, the first ten seconds of data are excluded in analysis. It has been observed a latency around $\SI{25}{\milli\second}$ for the camera. The image estimates have been shifted by the latency to be compared with the ground truth. For each pair of stereo images, the attitude is found by averaging the estimates from the two images. All the data are recorded and processed online.

\begin{figure}
  \centering
  \includegraphics[width=\linewidth]{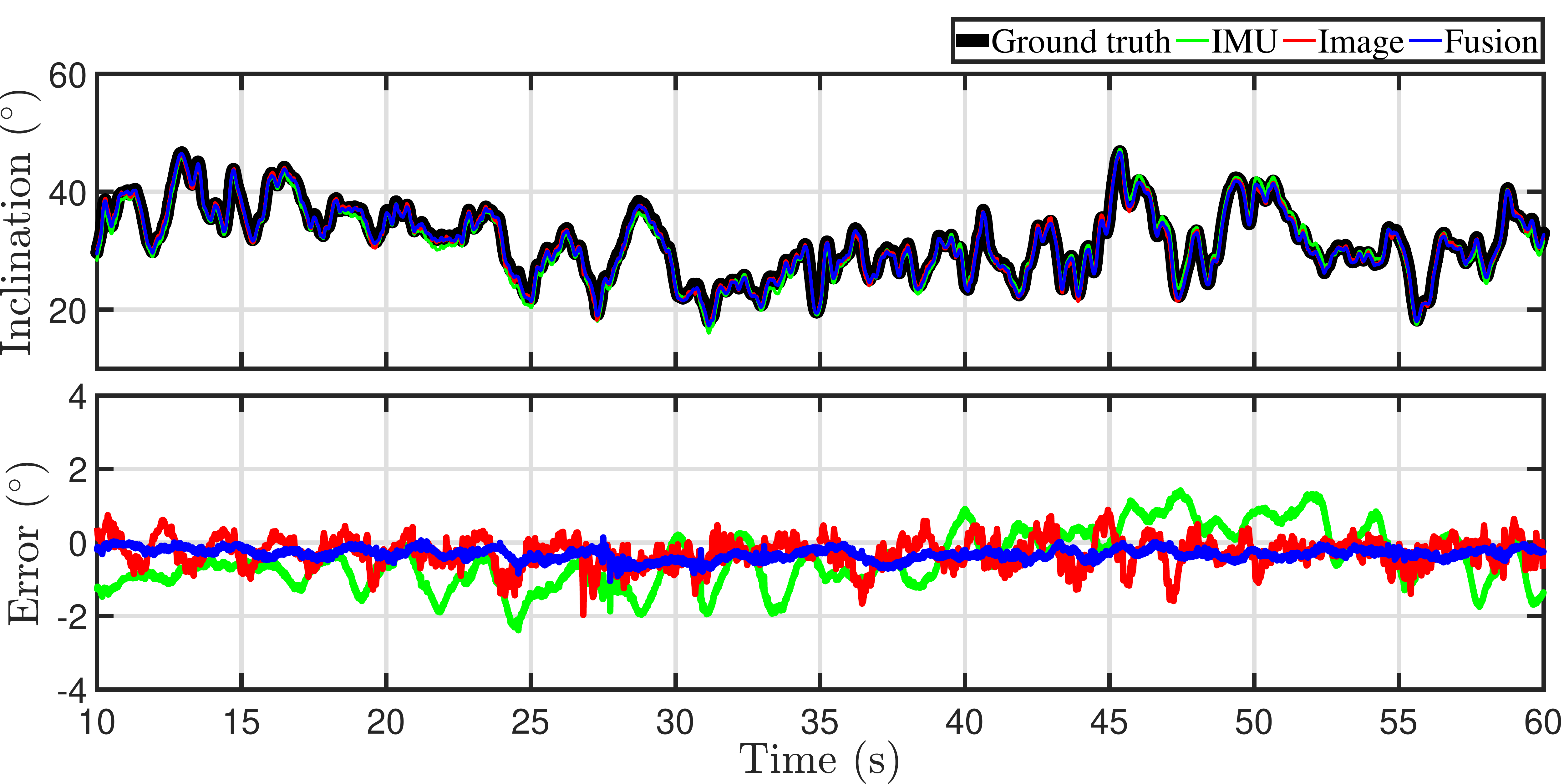}
  \caption{Inclination estimates from the Complementary Filter (IMU), 1080p videos and the fusion compared with the ground truth. }\label{1080dynamicResultWithError}
\end{figure}

\begin{table}
\renewcommand{\arraystretch}{1.3}
\caption{Inclination estimate errors of dynamic images}
\label{dynaResult}
\centering
\begin{tabular}{lcccc}
\hline\hline
  \multirow[l]{2}{*}{Resolution} & & \multicolumn{3}{c}{Inclination error ($^\circ$)}\\
  \hhline{~~-|-|-}
  & & IMU & Image & Fusion \\
\hline
  \multirow[l]{4}{*}{$1920\times1080$} & $\overline{\lvert e\rvert}$ & 0.850 & 0.376 & 0.301 \\
  & $\sigma$ & 0.967 & 0.403 & 0.155 \\
  & RMS & 1.036 & 0.484 & 0.338 \\
  & $f (\SI{}{\hertz})$ & 240 & 30 & 270 \\
\hline
  \multirow[l]{4}{*}{$1280\times720$} & $\overline{\lvert e\rvert}$ & 1.010& 0.525 & 0.267 \\
  & $\sigma$ & 0.949 & 0.710 & 0.284 \\
  & RMS & 1.274 & 0.733 & 0.351 \\
  & $f (\SI{}{\hertz})$ & 240 & 60 & 300 \\
\hline\hline
\end{tabular}
\end{table}

For 1080p videos, the estimates of inclination angles are shown in Fig. \ref{1080dynamicResultWithError}. The figure includes the ground truth from OptiTrack system, IMU estimates (the Complementary Filter), image estimates and estimates from the fusion. It is obvious from the figure that all estimates roughly overlap with the ground truth all along. Only IMU estimates occasionally have notable deviation from the ground truth. The bottom plot of Fig. \ref{1080dynamicResultWithError} shows the errors of each method. The numeric results of errors are listed in Table \ref{dynaResult}. It is clear that the image estimates and fusion estimates are notably more accurate than those from IMUs. It is noteworthy that the fusion has a significantly smaller standard deviation than other methods and an update frequency of $\SI{270}{\hertz}$ (with inconstant intervals) while images only update at $\SI{30}{\hertz}$. Another phenomenon in the figure is that the error of the Complimentary Filter usually reaches a local extremum when rotation changes abruptly. This is caused by the inherent drawback of the Complimentary Filter that it assumes the concerning object experiences no acceleration other than gravity. It is at the point when the pose changes fast, that assumption leads to the greatest errors. The proposed algorithm does not suffer from any of such assumptions and gives more stable result.

\begin{figure}
  \centering
  \includegraphics[width=\linewidth]{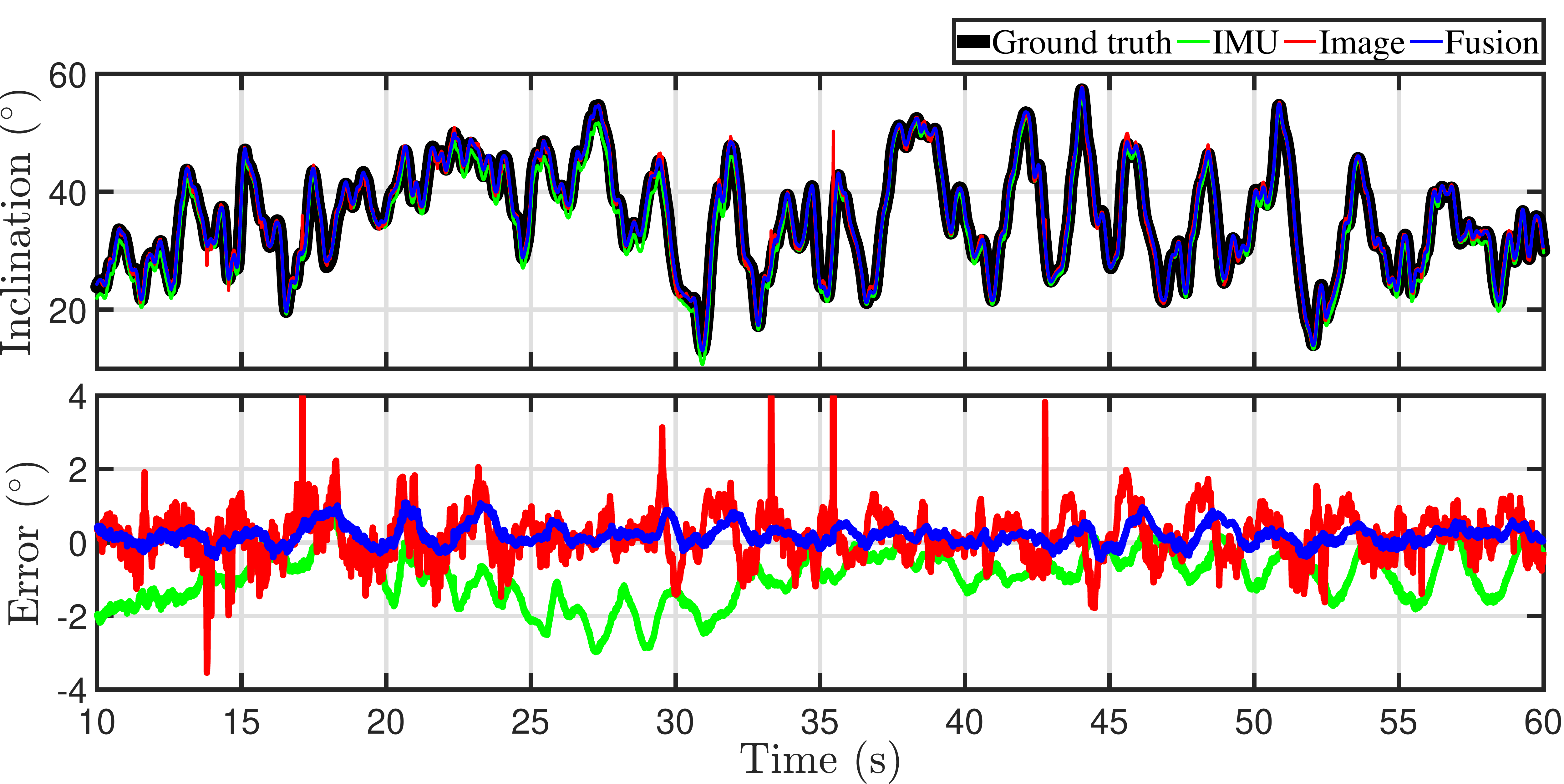}
  \caption{Inclination estimates from the Complementary Filter (IMU), 720p videos and the fusion compared with the ground truth. }\label{720dynamicResultWithError}
\end{figure}

For 720p videos, the estimates of inclination angles are presented in Fig. \ref{720dynamicResultWithError}. It has also been observed that all estimates roughly overlap with the ground truth all along. Only IMU estimates occasionally have notable deviation from the ground truth. The bottom plot in Fig. \ref{720dynamicResultWithError} shows the errors of each method. Compared with 1080p videos, the errors from image estimates have more sharp peaks. Further inspection into these peaks reveals that these errors are caused by noises in line detection. When a resolution reduces from 1080p to 720p, the number of edge pixels drops significantly. Thus, the line detection is more susceptible to outliers. The numerical results of errors are also listed in Table \ref{dynaResult}. Similar to the 1080p videos, the image estimates and fusion estimates are notably more accurate than those from IMUs. The fusion has a significantly smaller standard deviation and an update frequency of $\SI{300}{\hertz}$ (with inconstant intervals) while images only update at $\SI{60}{\hertz}$.

It is surprising that 720p videos yield estimates in the similar accuracy levels of 1080p videos after fusion, given that sole image estimates from 720p videos are relatively less accurate. The gyroscope in an IMU is generally very precise in a short term. The attitude estimates are deteriorated by noisy accelerometer measurements in the Complementary Filter. In fusion, image estimates completely replace the role of an accelerometer that to suppress drift in gyroscope estimates while maintaining high accuracy. Higher frequency of image estimates provides more opportunities to suppress the drift. This is likely one of the reasons why 720p videos give similarly accurate estimates after fusion. Nevertheless, the standard deviation of images estimates from the 720p video are still significantly larger.

In the two experiments, the errors from the conventional method drops significantly by more than $\SI{60}{\percent}$ in RMS errors after fusion. With regard to the standard deviation, the estimates become notably more stable after fusing image results. Even the estimates from solely images are evidently more accurate than the conventional method. The fusion combines the advantage of both methods that it has higher frequency and more accurate estimates. Considering that a 720p video can output 60 FPS in contrast to 30 FPS from a 1080p video and is much less computationally intensive, a resolution of $1280\times720$ is preferred for attitude estimation.

In practice, visual sensors have been hardly utilized for attitude estimation for robots. IMUs are much cheaper and easier to implement, and the accuracy of conventional method is sufficient in many scenarios. Vision sensors usually demand costly computation, thus suffering from low update frequency and long latency. However, the proposed method has processing time of only $\SI{1.41}{\milli\second}$ and an update frequency up to $\SI{60}{\hertz}$ for a 720p video or even much higher after fusion. The accuracy is also notably higher than the conventional method. Imagine a scenario when a UAV, such as a quadrotor, performs an aggressive maneuver to navigate through chokes or obstacles. An accurate and timely attitude estimate is the key to a precise control. Moreover, the conventional method suffers the most from the abrupt acceleration and deceleration where image estimates can play as a preferable complement.

\subsection{Position Estimation Result}

The same sets of data as previous attitude estimation are analyzed online for position estimation. As having been previously explained, the proposed method can locate the position of vertices across a pair of stereo images. Thus, the position of the vertex in the camera frame $\!{}^{\{\textnormal{C}\}}\!\bm{P}_\textnormal{v}$ can be found by triangulation. Since one already has the relative attitude $\!\bm{R}_\textnormal{a}$ in the object frame, the position of the camera in the object frame can be directly found by $\!{}^{\{\textnormal{O}\}}\!\bm{P}_\textnormal{c} = -\bm{R}_\textnormal{a}{}^{\{\textnormal{C}\}}\!\bm{P}_\textnormal{v}$.

\begin{figure*}
     \centering
     \begin{subfigure}[b]{0.5\linewidth}
         \centering
         \includegraphics[width=\linewidth]{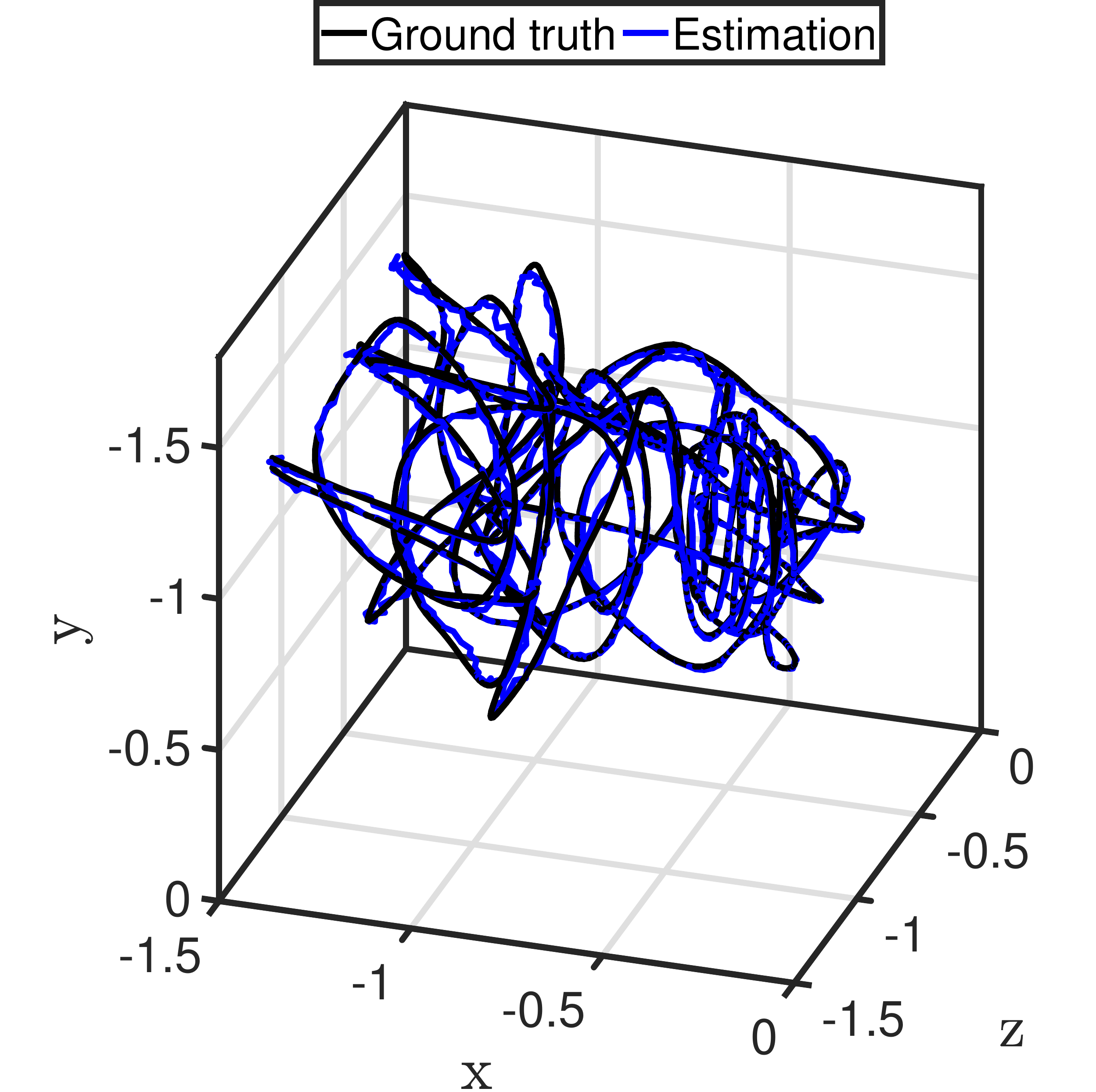}
         \caption{}
         \label{1080Traj}
     \end{subfigure}%
     \hfill
     \begin{subfigure}[b]{0.5\linewidth}
         \centering
         \includegraphics[width=\linewidth]{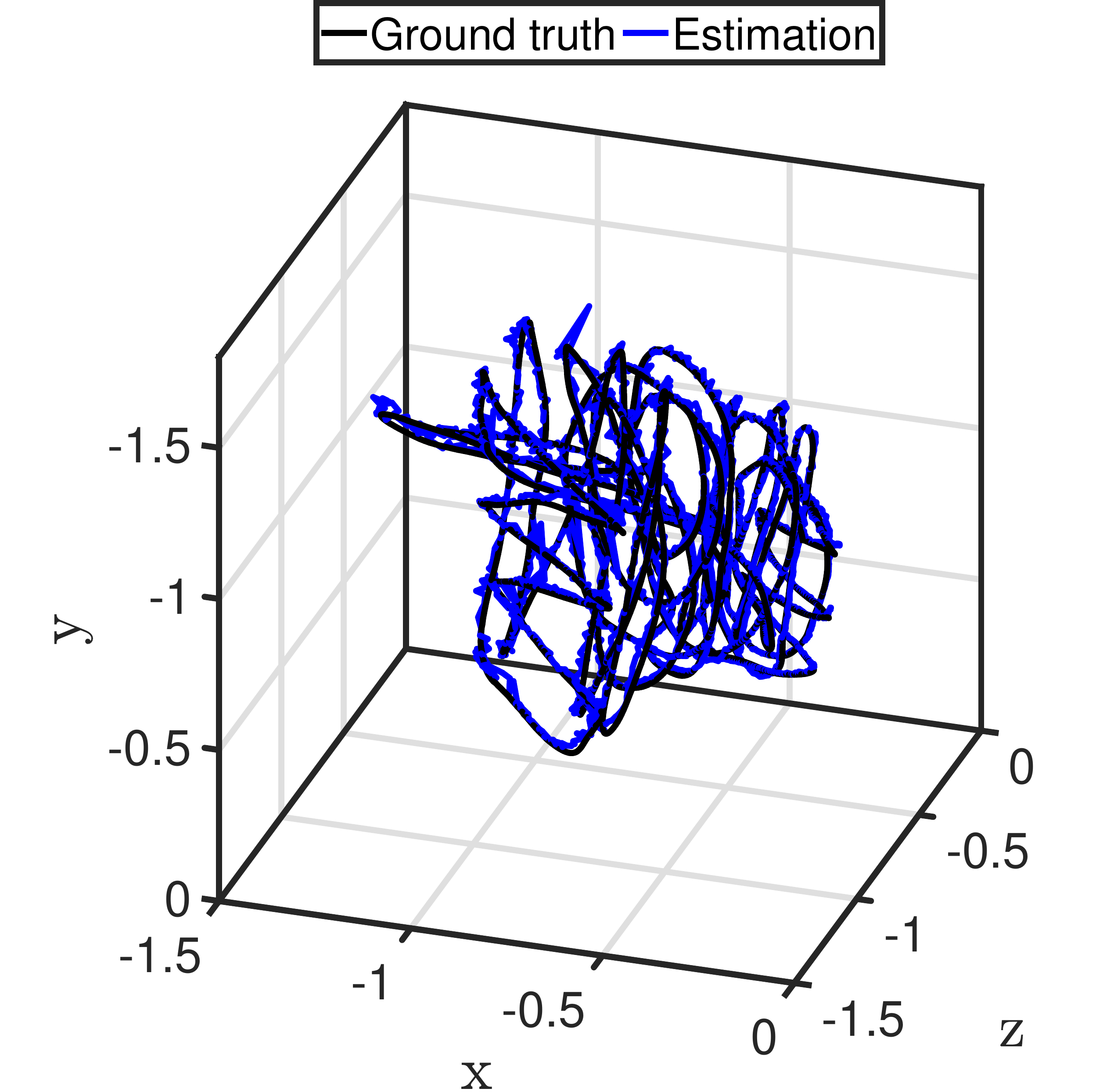}
         \caption{}
         \label{720Traj}
     \end{subfigure}%
     \caption{Estimated trajectory in the object frame. (a) from a 1080p video; (b) from a 720p video.}
     \label{allTraj}
\end{figure*}

The the ground truth and estimates of trajectories of the observer are displayed in Fig. \ref{allTraj}. It is noteworthy that the vertex of the object is at the origin and the x-, y- and z-component of the camera position is negative in the object frame. The trajectory estimated from a 1080p video smoothly follows the ground truth as shown in Fig. \ref{allTraj}(a). The trajectory estimated from a 720p video also follows the ground truth but with slightly more spikes, as shown in Fig. \ref{allTraj}(b). Similar to the errors in attitude estimation, the spikes are caused by noise in line detection. An incorrect edge line leads to deviation of the vertex and an incorrect attitude estimate, which both deteriorate position estimation.

\begin{figure}
\begin{subfigure}{\linewidth}
  \centering
  \includegraphics[width=\linewidth]{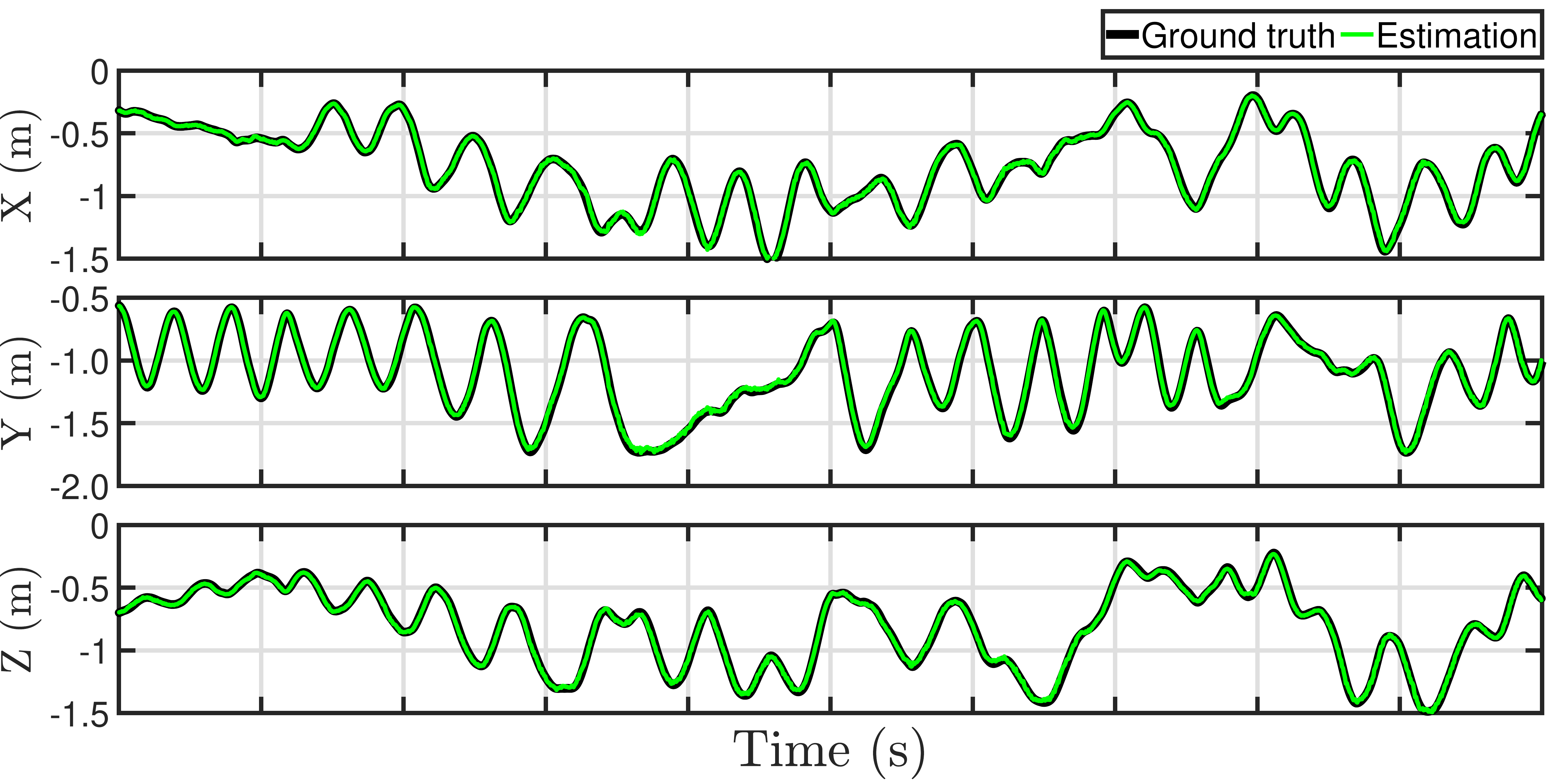}
  \caption{}\label{1080posiXYZ}
\end{subfigure}
\hfill
\begin{subfigure}{\linewidth}
  \centering
  \includegraphics[width=\linewidth]{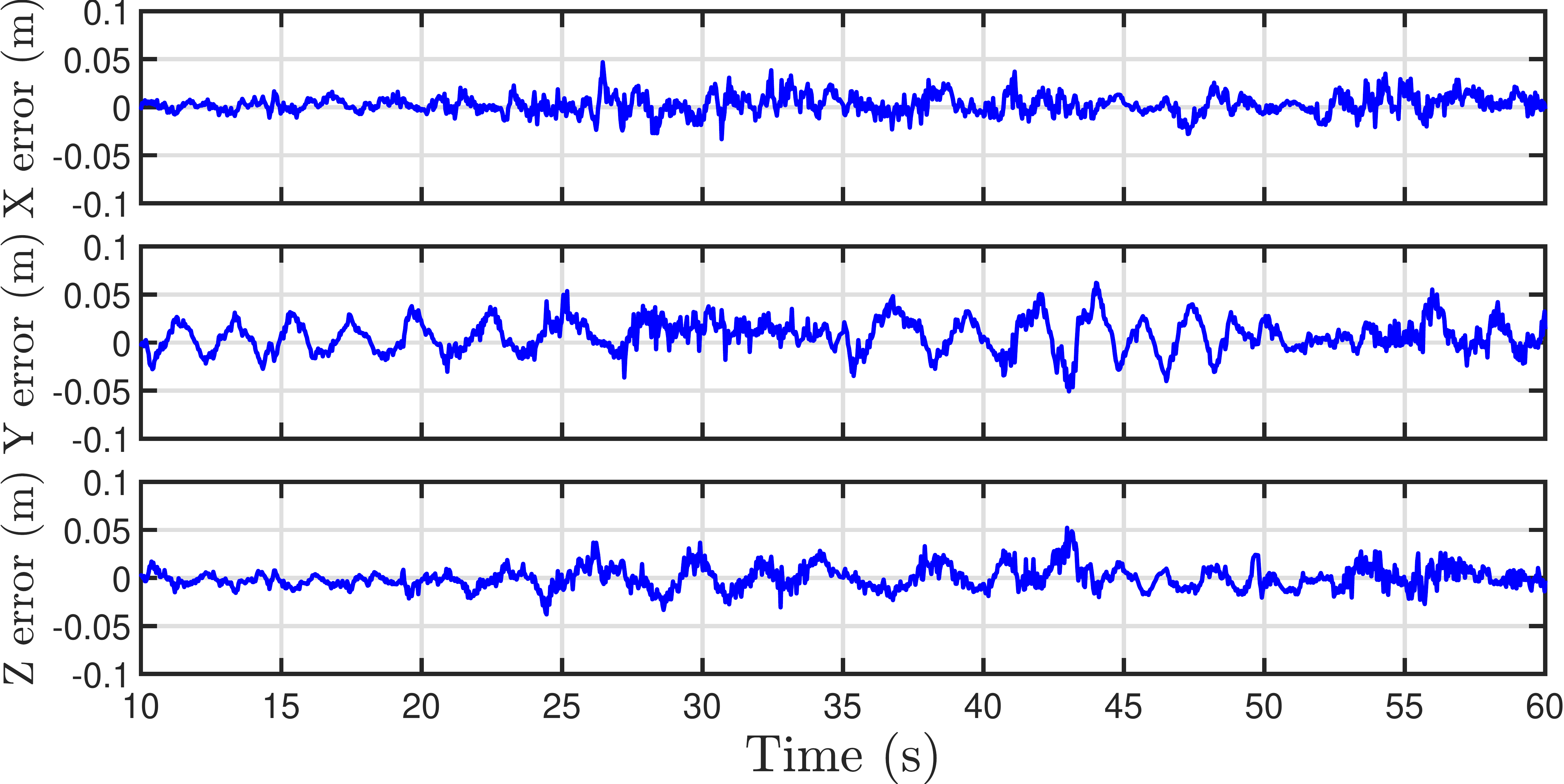}
  \caption{}\label{1080posiErrorXYZ}
\end{subfigure}
\caption{The trajectory estimated from a 1080p video. (a) the trajectory is displayed in the x-, y- and z- direction. The estimate follows close to the ground truth; (b) the trajectory error in x-, y- and z- direction. }\label{1080posi}
\end{figure}

\begin{figure}
\begin{subfigure}{\linewidth}
  \centering
  \includegraphics[width=\linewidth]{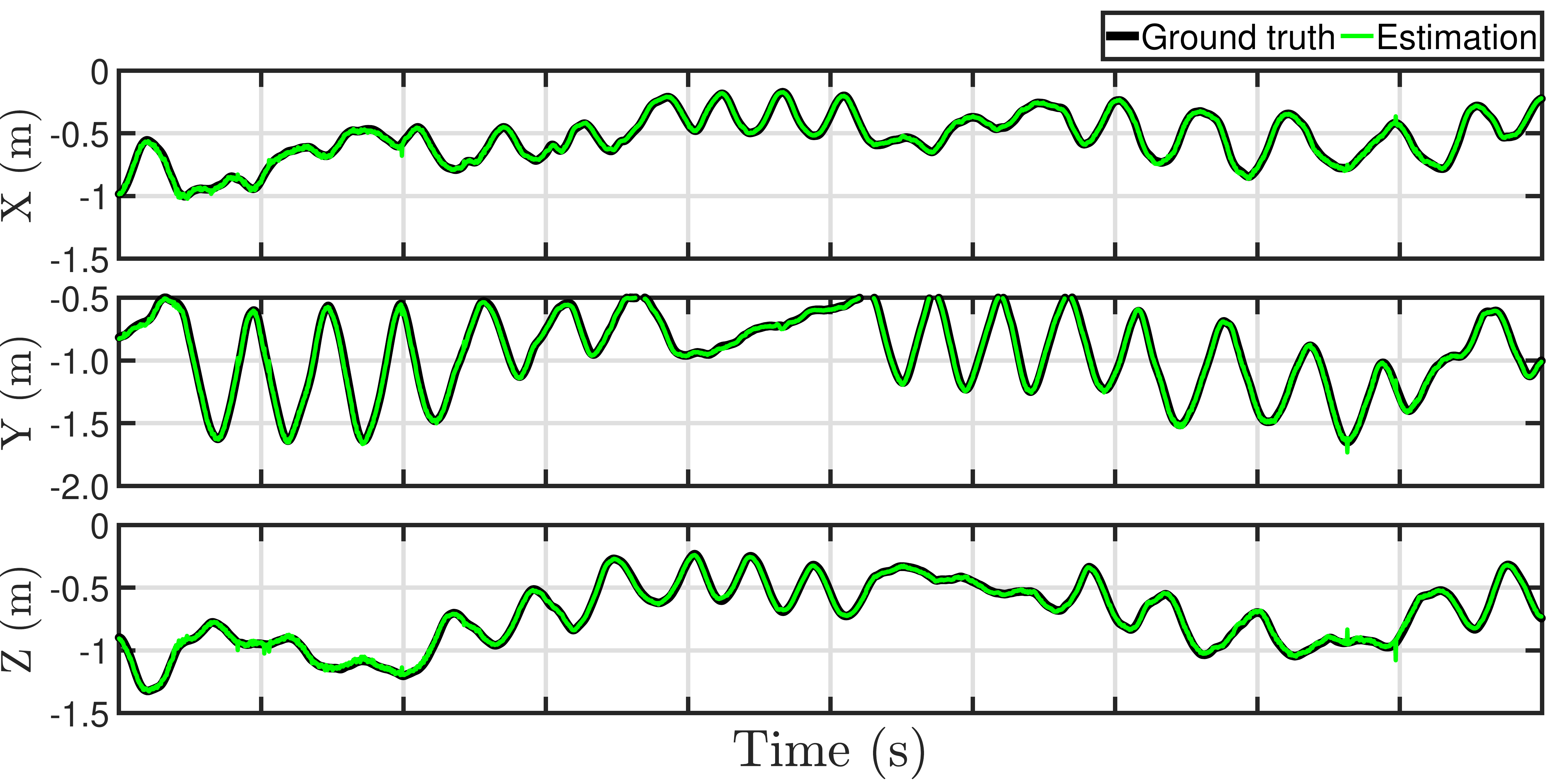}
  \caption{}\label{720posiXYZ}
\end{subfigure}
\hfill
\begin{subfigure}{\linewidth}
  \centering
  \includegraphics[width=\linewidth]{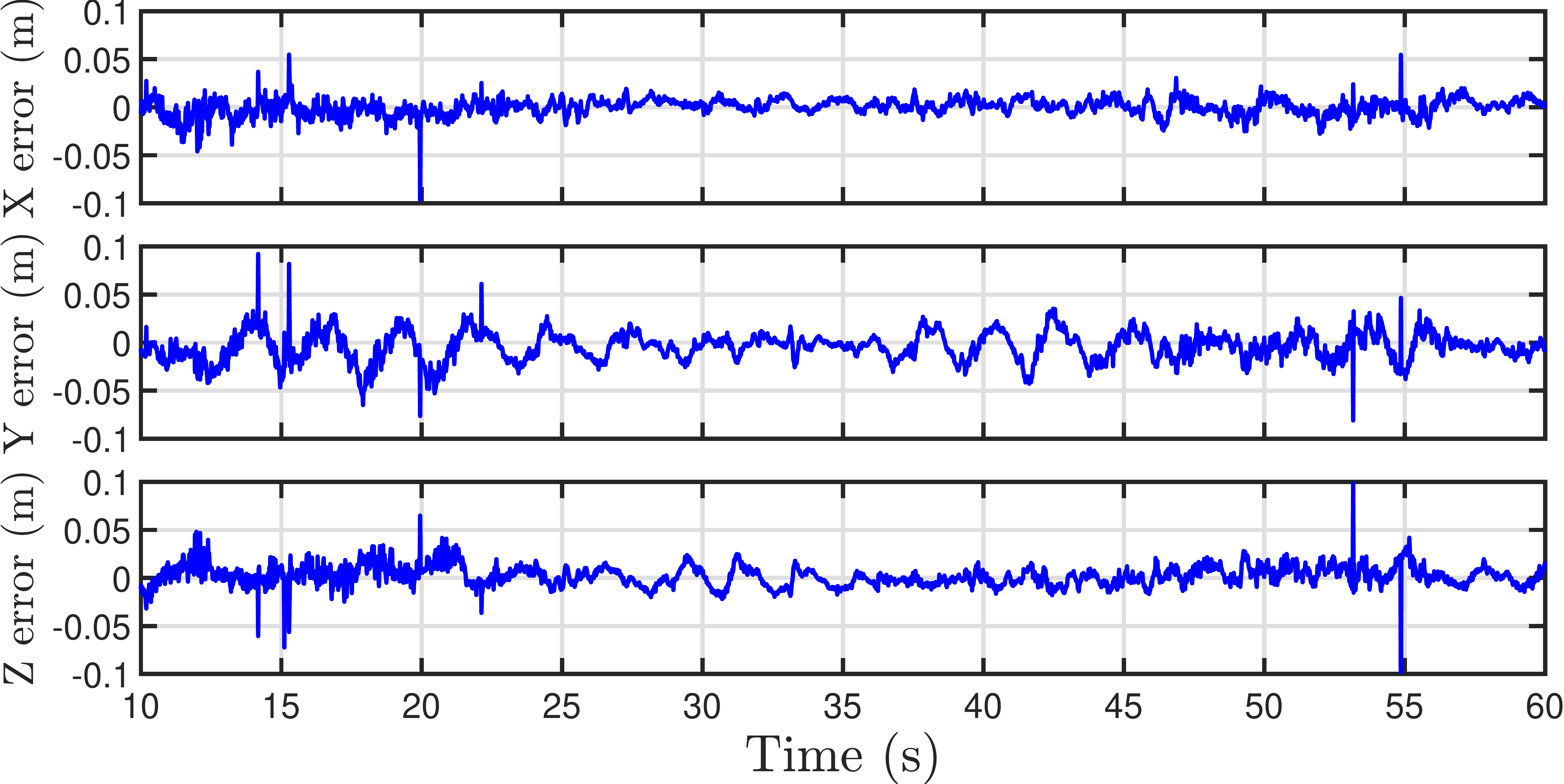}
  \caption{}\label{720posiErrorXYZ}
\end{subfigure}
\caption{The trajectory estimated from a 720 video. (a) the trajectory is displayed in the x-, y- and z- direction. The estimate follows close to the ground truth with slight spikes; (b) the trajectory error in x-, y- and z- direction. }\label{720posi}
\end{figure}

\begin{table}
\renewcommand{\arraystretch}{1.2}
\caption{Position estimate errors in Euler distance}
\label{Tab:posiErrorResult}
\centering
\begin{tabular}{ccccccc}
  \hline\hline
   & \multicolumn{6}{c}{Position error ($\SI{}{\centi\metre}$)}\\
  \hhline{~-|-|-|-|-|-}
   & \multicolumn{3}{c}{$1920\times1080$} & \multicolumn{3}{c}{$1280\times720$}\\
  \cmidrule(lr){2-4} \cmidrule(lr){5-7}
  Depth range (m) & $\overline{\lvert e\rvert}$ & $\sigma$ & RMS & $\overline{\lvert e\rvert}$ & $\sigma$ & RMS\\
  \hline
  $0.8-1.0$ & 1.22 &0.67 &1.39 &2.01 &1.03 &2.26\\
  $1.0-1.2$ & 1.69 &0.85 &1.89 &2.27 &1.17 &2.55\\
  $1.2-1.4$ & 1.99 &1.17 &2.31 &2.23 &1.57 &2.73\\
  $1.4-1.6$ & 2.17 &0.86 &2.33 &2.49 &1.27 &2.79\\
  $1.6-1.8$ & 2.45 &1.14 &2.71 &3.05 &1.68 &3.48\\
  $1.8-2.0$ & 2.69 &1.00 &2.87 &2.93 &1.84 &3.46\\
  $2.0-2.2$ & 2.27 &1.05 &2.50 &2.99 &1.41 &3.31\\
  \hline
  Average & 2.07 & 0.96 & 2.28 & 2.57 & 1.42 & 2.94\\
  \hline\hline
\end{tabular}
\end{table}

The trajectories and corresponding errors in x-, y- and z-direction of videos at resolutions of $1920\times1080$ and $1280\times720$ are displayed in Figs. \ref{1080posi} and \ref{720posi}, respectively. Both estimates follow close to the ground truth. The estimates from the 1080p video are more smooth, whereas occasional sharp errors present in estimates from the 720p video. For binocular stereo vision, the depth of an object $L$ is determined by $Bf/d$ where $B$ is the stereo baseline, $d$ is the disparity and $f$ is the focal length. The depth of an object is the z-component of the object position $(X,Y,Z)$ in the camera frame. By derivation and transformation, one has $\partial{Z}/\partial{d}=-L^2/(Bf)$. Similarly, one can determine that $\partial{X}/\partial{d}=-uL^2/(Bf^2)$ and $\partial{Y}/\partial{d}=-vL^2/(Bf^2)$, where $(u,v)$ is the position of the vertex in the image. The equation indicates that the depth error and a part of $X$ and $Y$ error are approximately proportional to the square of disparity. Therefore, the errors in Euler distance are listed against depth as shown in Table \ref{Tab:posiErrorResult}. The results in the table suggest that estimates from the 1080p video are better in all the average error magnitude, standard deviation and RMS. Considering the fact that a 720p video can update at $\SI{60}{\hertz}$ compared with $\SI{30}{\hertz}$ of a 1080p video without compromising too much on accuracy, 720p videos are more preferable. It is also the common practice in SLAM implementation that frequency weighs more than accuracy.

\begin{figure}
  \centering
  \includegraphics[width=\linewidth]{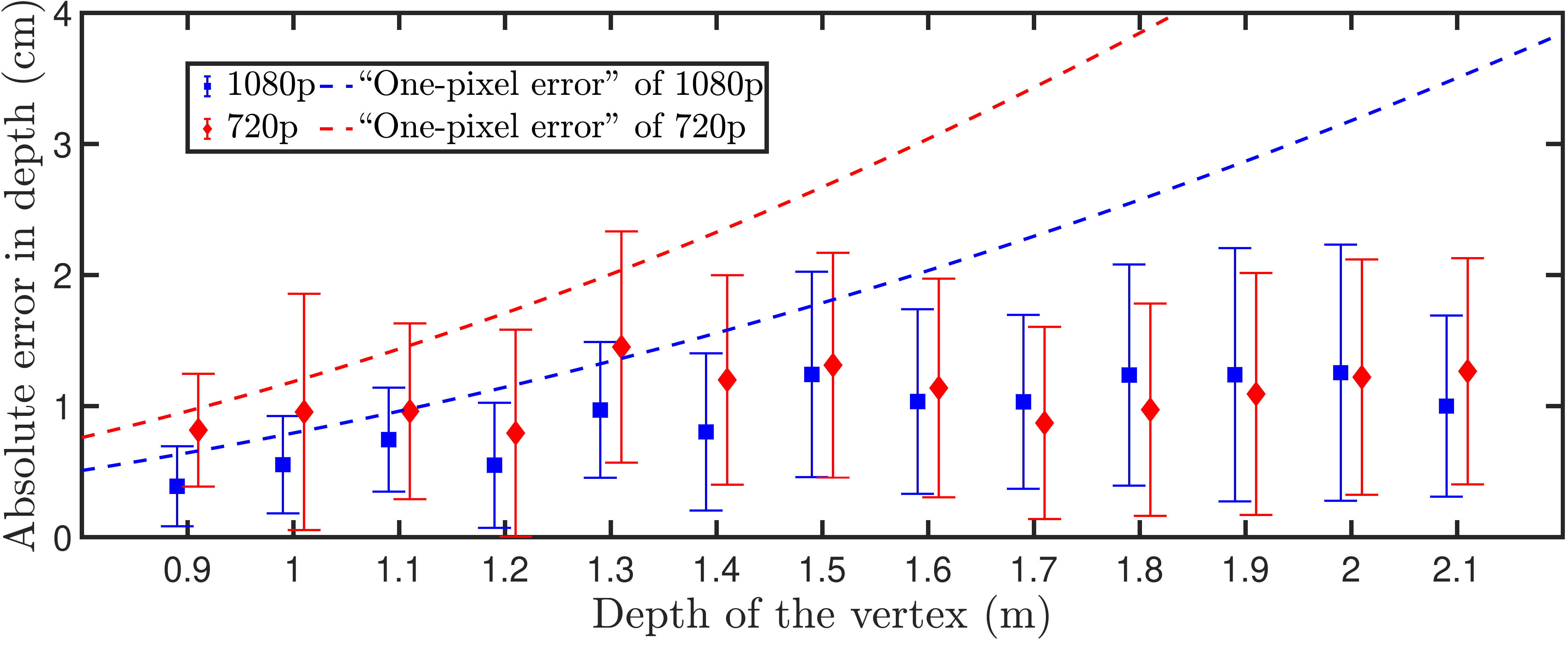}
  \caption{The absolute error of depth is plotted versus the depth of the vertex from 1080p videos (light blue) and 720p videos (blue). The ``one-pixel'' error is predicted by assuming an error of one pixel in disparity.}\label{depthErrorBarComparison}
\end{figure}

The average and standard deviation of absolute depth errors is investigated and presented in Fig. \ref{depthErrorBarComparison}. A number on the x-axis represents the range $\pm0.05$ to itself, for example $0.9$ represents the range of $(0.85,0.95)$. The figure indicates that the accuracy of depth is not strongly affected by resolutions. 1080p videos have obviously better accuracy over 720p videos only near the object. By assuming an error in disparity of one pixel, one can approximate corresponding depth errors by the equation of $e_Z=e_dL^2/(Bf)$. The approximated depth errors for 1080p and 720p videos are also displayed in the figure, labeled as ``one-pixel error''. As shown in the figure, all the average error magnitude is notably below the ``one-pixel error'' and the actual error of depth does not grow quadratically. The sudden rise for 720p videos near the depth of $\SI{1.3}{\metre}$ is caused by some incidental noise in line detection. In terms of disparity errors by $e_Z=e_dL^2/(Bf)$, the further the camera the better the accuracy generally becomes. For example, the average disparity error for 720p videos is estimated to be 0.85 pixels at the depth of $\SI{0.9}{\metre}$ and 0.36 pixels at the depth of $\SI{2.1}{\metre}$.

\section{Conclusion and Future Work}

This paper proposed a novel feature, named as ``TROVE feature''. By observing the feature using a color stereo camera in an environment that is rich in man-made structures, 6-DoF ego-states can be efficiently obtained in high accuracy. A TROVE feature consists of three rays and one vertex in an image. The feature is projected from a common structure, named as ``TROVE structure'', on man-made buildings and objects. A TROVE structure consists of three edges, one vertical and two horizontal, intersecting at one vertex. These edges are usually the dominating edges in an indoor environment, with the vertex being the corner of a building, man-made objects and etc. Such a structure favors navigation as a stationary reference. Many characteristics of a TROVE feature are associated with the physical properties of the corresponding TROVE structure. Compared with many commonly used features as SIFT, SURF, FAST, LSD, ORB/ORB2 and etc., the TROVE feature is more rich in physical information. This can possibly give more clues when conducting tasks on data registration. The angle between the two horizontal edges is known as priori knowledge and can be any angle between $0$ to $\pi$. By imagining a parallelepiped out of a TROVE structure, attitudes and positions can be obtained regarding the frame uniquely defined by the parallelepiped. Definition and how such an imaginary parallelepiped can be constructed is given in detail. It is proved in this paper that only two possible configurations of the imaginary parallelepiped can be constructed from an image of the TROVE feature. In most cases, the solution is proved as unique. Easy-to-implement solutions are given to discard incorrect interpretations and solutions. The algorithm is very efficient that for an image at $1920\times1080$ the processing time is merely $\SI{3.267}{\milli\second}$ and for an image at $1280\times720$ the processing time is only $\SI{1.411}{\milli\second}$.

Experiments have been conducted to evaluate the effectiveness and errors in pose estimation. Compared with the conventional method utilizing IMUs, the proposed method has significant improvement in accuracy without any drift. By fusing image estimates, the update frequency has reached $\SI{300}{\hertz}$ while RMS error drops by more than $\SI{60}{\percent}$ compared with the conventional method. It has also been discovered that different resolution can result in similar accuracy after the fusion. Although giving less accurate estimates, lower-resolution images can update much more frequently. More importantly, localization can also be achieved by the proposed algorithm. The RMS error in position is relatively small as $\SI{2.22}{\centi\metre}$ and $\SI{2.82}{\centi\metre}$ respectively for a 1080p video and a 720p video. In the context of indoor environment, the accuracy is sufficient for most unmanned robots to navigate. Note that a 720p video offers estimates at $\SI{60}{\hertz}$ and consumes much less computation without compromising too much on accuracy. Thus, it is more desirable and efficient for implementations especially for robots that have limited payload capability such as UAVs.

As a preliminary work to evaluate the feasibility and accuracy of the proposed method, the structure to be observed is represented by a colored object in the experiment. Future work will extend the approach to more general cases, where actual indoor scenes would be captured for analysis. In the actual implementation, losing sight of any TROVE structures is a common challenge the method will encounter. An active gimbal will be mounted to lock the camera facing to the structure.

\section*{Acknowledgment}

The authors would like to thank Carlos Ma, Ka Wah Lee and Wenyutian Xu for their kind help in setting up and conducting the experiments.

\ifCLASSOPTIONcaptionsoff
  \newpage
\fi

\bibliographystyle{IEEEtran}
\bibliography{RefVisionOdometry_ArXiv}

\begin{thebibliography}{10}
\providecommand{\url}[1]{#1}
\csname url@samestyle\endcsname
\providecommand{\newblock}{\relax}
\providecommand{\bibinfo}[2]{#2}
\providecommand{\BIBentrySTDinterwordspacing}{\spaceskip=0pt\relax}
\providecommand{\BIBentryALTinterwordstretchfactor}{4}
\providecommand{\BIBentryALTinterwordspacing}{\spaceskip=\fontdimen2\font plus
\BIBentryALTinterwordstretchfactor\fontdimen3\font minus
  \fontdimen4\font\relax}
\providecommand{\BIBforeignlanguage}[2]{{%
\expandafter\ifx\csname l@#1\endcsname\relax
\typeout{** WARNING: IEEEtran.bst: No hyphenation pattern has been}%
\typeout{** loaded for the language `#1'. Using the pattern for}%
\typeout{** the default language instead.}%
\else
\language=\csname l@#1\endcsname
\fi
#2}}
\providecommand{\BIBdecl}{\relax}
\BIBdecl

\bibitem{engel2015large}
J.~Engel, J.~St{\"u}ckler, and D.~Cremers, ``Large-scale direct {SLAM} with
  stereo cameras,'' in \emph{2015 IEEE/RSJ IEEE/RSJ International Conference on
  Intelligent Robots and Systems}, 2015, pp. 1935--1942.

\bibitem{mur2017orb}
R.~Mur-Artal and J.~D. Tard{\'o}s, ``{ORB-SLAM}2: An open-source {SLAM} system
  for monocular, stereo, and {RGB-D} cameras,'' \emph{IEEE Transactions on
  Robotics}, vol.~33, no.~5, pp. 1255--1262, 2017.

\bibitem{pire2017s}
T.~Pire, T.~Fischer, G.~Castro, P.~De~Crist{\'o}foris, J.~Civera, and J.~J.
  Berlles, ``{S-PTAM}: Stereo parallel tracking and mapping,'' \emph{Robotics
  and Autonomous Systems}, vol.~93, pp. 27--42, 2017.

\bibitem{lowe2004distinctive}
D.~G. Lowe, ``Distinctive image features from scale-invariant keypoints,''
  \emph{International Journal of Computer Vision}, vol.~60, no.~2, pp. 91--110,
  2004.

\bibitem{bay2006surf}
H.~Bay, T.~Tuytelaars, and L.~Van~Gool, ``{SURF}: Speeded up robust features,''
  in \emph{European Conference on Computer Vision}, 2006, pp. 404--417.

\bibitem{rosten2006machine}
E.~Rosten and T.~Drummond, ``Machine learning for high-speed corner
  detection,'' in \emph{European Conference on Computer Vision}, 2006, pp.
  430--443.

\bibitem{horn1988closed}
B.~K. Horn, H.~M. Hilden, and S.~Negahdaripour, ``Closed-form solution of
  absolute orientation using orthonormal matrices,'' \emph{Journal of the
  Optical Society of America A}, vol.~5, no.~7, pp. 1127--1135, 1988.

\bibitem{haralick1991analysis}
R.~M. Haralick, D.~Lee, K.~Ottenburg, and M.~Nolle, ``Analysis and solutions of
  the three point perspective pose estimation problem,'' in \emph{Proceedings
  of the 1991 IEEE Computer Society Conference on Computer Vision and Pattern
  Recognition}, 1991, pp. 592--598.

\bibitem{lepetit2009epnp}
V.~Lepetit, F.~Moreno-Noguer, and P.~Fua, ``{EPnP}: An accurate {O(n)} solution
  to the {PnP} problem,'' \emph{International Journal of Computer Vision},
  vol.~81, no.~2, p. 155, 2009.

\bibitem{claus2005reliable}
D.~Claus and A.~W. Fitzgibbon, ``Reliable automatic calibration of a
  marker-based position tracking system,'' in \emph{the 7th IEEE Workshop on
  Applications of Computer Vision}, vol.~1, 2005, pp. 300--305.

\bibitem{fiala2010designing}
M.~Fiala, ``Designing highly reliable fiducial markers,'' \emph{IEEE
  Transactions on Pattern Analysis and Machine Intelligence}, vol.~32, no.~7,
  pp. 1317--1324, 2010.

\bibitem{bergamasco2016accurate}
F.~Bergamasco, A.~Albarelli, L.~Cosmo, E.~Rodola, and A.~Torsello, ``An
  accurate and robust artificial marker based on cyclic codes.'' \emph{IEEE
  Transactions on Pattern Analysis and Machine Intelligence}, vol.~38, no.~12,
  pp. 2359--2373, 2016.

\bibitem{munoz2018mapping}
R.~Mu{\~n}oz-Salinas, M.~J. Mar{\'\i}n-Jimenez, E.~Yeguas-Bolivar, and
  R.~Medina-Carnicer, ``Mapping and localization from planar markers,''
  \emph{Pattern Recognition}, vol.~73, pp. 158--171, 2018.

\bibitem{coughlan1999manhattan}
J.~M. Coughlan and A.~L. Yuille, ``Manhattan world: Compass direction from a
  single image by {Bayesian} inference,'' in \emph{Proceedings of the 7th IEEE
  International Conference on Computer Vision}, vol.~2, 1999, pp. 941--947.

\bibitem{denis2008efficient}
P.~Denis, J.~H. Elder, and F.~J. Estrada, ``Efficient edge-based methods for
  estimating manhattan frames in urban imagery,'' in \emph{European Conference
  on Computer Vision}, 2008, pp. 197--210.

\bibitem{furukawa2009manhattan}
Y.~Furukawa, B.~Curless, S.~M. Seitz, and R.~Szeliski, ``Manhattan-world
  stereo,'' in \emph{2009 IEEE Conference on Computer Vision and Pattern
  Recognition}, 2009, pp. 1422--1429.

\bibitem{dusha2007attitude}
D.~Dusha, W.~Boles, and R.~Walker, ``Attitude estimation for a fixed-wing
  aircraft using horizon detection and optical flow,'' in \emph{the 9th
  Biennial Conference of the Australian Pattern Recognition Society on Digital
  Image Computing Techniques and Applications}, 2007, pp. 485--492.

\bibitem{demonceaux2007uav}
C.~Demonceaux, P.~Vasseur, and C.~P{\'e}gard, ``{UAV} attitude computation by
  omnidirectional vision in urban environment,'' in \emph{Proceedings of the
  2007 IEEE International Conference on Robotics and Automation}, 2007, pp.
  2017--2022.

\bibitem{shabayek2012vision}
A.~E.~R. Shabayek, C.~Demonceaux, O.~Morel, and D.~Fofi, ``Vision based {UAV}
  attitude estimation: Progress and insights,'' \emph{Journal of Intelligent \&
  Robotic Systems}, vol.~65, no.~1, pp. 295--308, 2012.

\bibitem{hartley2003multiple}
R.~Hartley and A.~Zisserman, \emph{Multiple View Geometry in Computer
  Vision}.\hskip 1em plus 0.5em minus 0.4em\relax Cambridge University Press,
  2003.

\bibitem{dhome1989determination}
M.~Dhome, M.~Richetin, J.-T. Lapreste, and G.~Rives, ``Determination of the
  attitude of {3D} objects from a single perspective view,'' \emph{IEEE
  Transactions on Pattern Analysis and Machine Intelligence}, vol.~11, no.~12,
  pp. 1265--1278, 1989.

\bibitem{bazin2012globally}
J.-C. Bazin, Y.~Seo, C.~Demonceaux, P.~Vasseur, K.~Ikeuchi, I.~Kweon, and
  M.~Pollefeys, ``Globally optimal line clustering and vanishing point
  estimation in {Manhattan} world,'' in \emph{2012 IEEE Conference on Computer
  Vision and Pattern Recognition}, 2012, pp. 638--645.

\bibitem{kurugollu2001color}
F.~Kurugollu, B.~Sankur, and A.~E. Harmanci, ``Color image segmentation using
  histogram multithresholding and fusion,'' \emph{Image and Vision Computing},
  vol.~19, no.~13, pp. 915--928, 2001.

\bibitem{liu2017novel}
Y.~Liu and M.~Z.~Q. Chen, ``A novel three-axis visual attitude estimation
  algorithm with application to quadrotors,'' in \emph{the 29th Chinese Control
  Decision Conference}, 2017, pp. 5436--5441.

\bibitem{fischler1981random}
M.~A. Fischler and R.~C. Bolles, ``Random sample consensus: a paradigm for
  model fitting with applications to image analysis and automated
  cartography,'' \emph{Communications of the ACM}, vol.~24, no.~6, pp.
  381--395, 1981.

\bibitem{weng1992camera}
J.~Weng, P.~Cohen, and M.~Herniou, ``Camera calibration with distortion models
  and accuracy evaluation,'' \emph{IEEE Transactions on Pattern Analysis and
  Machine Intelligence}, vol.~14, no.~10, pp. 965--980, 1992.

\bibitem{corke2011robotics}
P.~Corke, \emph{Robotics, Vision and Control: Fundamental Algorithms in
  MATLAB}.\hskip 1em plus 0.5em minus 0.4em\relax Springer, 2011, vol.~73,
  ch.~10, pp. 251--283.

\bibitem{mahony2005complementary}
R.~Mahony, T.~Hamel, and J.-M. Pflimlin, ``Complementary filter design on the
  special orthogonal group {SO(3)},'' in \emph{the 44th IEEE Conf. Decision
  Control and 2005 European Control Conf.}, 2005, pp. 1477--1484.

\bibitem{dam1998quaternions}
E.~B. Dam, M.~Koch, and M.~Lillholm, \emph{Quaternions, Interpolation and
  Animation}.\hskip 1em plus 0.5em minus 0.4em\relax Citeseer, 1998, vol.~2.

\end{thebibliography}

\end{document}